\documentclass{article}
\usepackage{arxiv}

\usepackage{stmaryrd}

\newtheorem{theorem}{Theorem}%  meant for continuous numbers
\newtheorem{proposition}{Proposition}% to get separate numbers for theorem and proposition etc.
\newtheorem{corollary}{Corollary}%
\newtheorem{definition}{Definition}%

\usepackage{subcaption}

\newtheorem{lemma}{Lemma}
\usepackage{stmaryrd}
\usepackage{multirow}
\usepackage[table,xcdraw]{xcolor}
\newcommand{\ro}[1]{{\color{black}{#1}}}
\newcommand{\ma}[1]{{\color{black}{#1}}}
\newcommand{\edu}[1]{{\color{black}{#1}}}

\newcommand{\Dgm}{\mbox{Dgm}}
\usepackage{multicol}
\usepackage{multirow}
\usepackage{mathrsfs}
\usepackage{amsmath,accents}
\usepackage{algpseudocode}
\usepackage{algorithm}
\usepackage{hyperref}
\newcommand{\N}{{\mathbb N}}
\newcommand{\cN}{{\cal N}}

\usepackage{amsfonts}
\usepackage[utf8]{inputenc} 
\usepackage{pgf,tikz}
\usepackage{program}
\usepackage{natbib}
\usepackage{makecell, caption}
\usepackage{array}
    \newcolumntype{P}[1]{>{\centering\arraybackslash}p{#1}}
    \newcolumntype{M}[1]{>{\centering\arraybackslash}m{#1}}

\hypersetup{
pdftitle={Topology-based
Representative Datasets 
to Reduce Neural Network Training Resources},
pdfauthor={Rocio Gonzalez-Diaz et.al},
pdfkeywords={Data Reduction, Neural Networks, Representative Datasets, Computational Topology}
}

\begin{document}

\title{Topology-based
Representative Datasets 
to Reduce Neural Network Training Resources}


\author{ Rocio Gonzalez-Diaz \\
    Applied Math I department \\
    University of Sevilla\\
    rogodi@us.es
	%% examples of more authors
	\And
	Miguel A. Gutiérrez-Naranjo\\
	Computer Sciences and Artificial Intelligence department \\
    University of Sevilla\\
    magutier@us.es
	\AND
	Eduardo Paluzo-Hidalgo\\
	Applied Math I department \\
    University of Sevilla\\
    epaluzo@us.es\thanks{Corresponding author.}
	%% \AND
	%% Coauthor \\
	%% Affiliation \\
	%% Address \\
	%% \texttt{email} \\
	%% \And
	%% Coauthor \\
	%% Affiliation \\
	%% Address \\
	%% \texttt{email} \\
	%% \And
	%% Coauthor \\
	%% Affiliation \\
	%% Address \\
	%% \texttt{email} \\
}

\maketitle

\begin{abstract}
One of the main drawbacks of the practical use of neural networks is the long time required in the training process. Such a training process consists of an iterative change of parameters trying to minimize a loss function. These changes are driven by a dataset, which can be seen as a set of labelled points in an n-dimensional space. In this paper, we explore the concept of a {\it representative dataset} which is a dataset smaller than the original one, 
%and which satisfies 
\ro{satisfying} a nearness condition independent of isometric transformations. Representativeness is measured using persistence diagrams %(%which is 
(a computational topology tool) due to its computational efficiency. We prove that the accuracy of the learning process of a neural network on a representative dataset is ``similar'' to the accuracy on the original dataset when the neural network architecture is a perceptron and the loss function is the mean squared error. These theoretical results accompanied by experimentation open a door to 
reducing the size of the dataset to gain time in the training process of any neural network.
\end{abstract}

\section{Introduction}\label{sec:introduction}
The success of the different architectures used in the framework of neural networks is
doubtless \cite{Goodfellow-et-al-2016}. 
The achievements made in areas such as video imaging
\cite{DBLP:journals/corr/abs-2103-01739},
recognition 
\cite{DBLP:journals/corr/abs-2108-11821}, or 
language models \cite{DBLP:journals/corr/abs-2005-14165}
show the surprising potential of such architectures. In spite of such success, they still have some shortcomings.
One of their main drawbacks is the long time needed in the training process. Such a long training time is usually associated with two factors: first, the large amount of weights to be adjusted in the current architectures and, second, the huge datasets used to train neural networks. 
In general, the time needed to train
a complex neural network from the scratch is so long that many researchers use {\it pretrained} neural networks as,
for example,
Oxford VGG models \cite{SimonyanZ2014}, Google Inception Model \cite{SzegedyLJSRAEVR2015}, or Microsoft ResNet Model \cite{1512.03385}.
Other attempts  to reduce the training time  are, for example, to partition the training task in multiple training subtasks with submodels, which can be performed independently and in parallel
\cite{1511.02954}, to use asynchronous averaged stochastic gradient descent \cite{DBLP:conf/iscslp/YouX14}, and to reduce  data  transmission through a sampling-based approach \cite{DBLP:conf/ijcai/XiaoZLZ17}. \edu{Besides, in \cite{8576015}, the authors \ro{studied}
%study 
how \ro{the elimination of ``unfavourable''}
%dropping unfavourable 
samples \ro{improves} 
%the 
generalization \ro{accuracy} for convolutional neural networks.
\ro{Finally,}
%and 
in \cite{Wang_Zhu_Dong_He_Huang_2020}, an unweighted influence data subsampling method is proposed.}

Roughly speaking, a training process consists of searching for a local minimum of a loss function in an abstract space where the {\it states} are sets of weights. 
Each of the training sample batches provides an extremely small change in weights according to the training rules. 
The aim of such changes is to find the ``best'' set of weights that minimizes the loss function.
If we consider a ``geometrical'' interpretation of the learning process, such changes can be seen as tiny steps in a multidimensional metric space of parameters that follow the direction settled by the gradient of the loss function. 
In some sense, one may think that two ``close'' points of the dataset with the same 
label provide ``similar''  information to the learning process since the gradient of the loss function on such points is  similar. 

Such a viewpoint leads us to look for a 
{\it representative dataset}, ``close'' to and with a  smaller number of points than the original dataset but keeping its ``topological information'', allowing the neural network to perform the learning process taking less time without losing accuracy.
This way, in this paper, we show how to  reduce the training time by choosing a
representative dataset  of the original dataset.

Besides, we formally prove, for the perceptron case, that the accuracy of a neural network trained with the representative dataset is 
\ro{similar}
%comparable 
to the accuracy of the neural network trained with the original dataset.
Experimental evidence indicates 
\ro{the same property in}
%similar behaviour of 
representative datasets  for the general case.

Moreover, in order to ``keep the shape'' of the original dataset, the concept of representative datasets is associated with a notion of nearness independent of  isometric transformations. 
As a first approach, the Gromov-Hausdorff distance is used  to measure the
%this
{\it representativeness} of the dataset. Nonetheless, as the Gromov-Hausdorff distance complexity is an open problem\footnote{It seems to be intractable in practice \cite{DBLP:journals/dcg/Schmiedl17}.}, the bottleneck distance between persistence diagrams \cite{Edelsbrunner10} is used instead as a lower bound to the Gromov-Hausdorff distance since its time complexity is cubic on the size of the dataset (see \cite{chazalsignature}).

The paper is organized as follows. In Section \ref{sec:background}, basic definitions and results from neural networks and computational topology are given. The notion of representative datasets is introduced in Section \ref{sec:repdatasets}. 
Persistence diagrams are used in Section \ref{section:TDA} to measure the representativeness of a dataset. In Section \ref{sec:perceptroncase}, the perceptron architecture is studied to provide bounds for comparing training performances within the original dataset and its representative dataset. 
Specifically, in Subsection \ref{sec:experimentalresults}, experimental results are provided for the perceptron case showing the good performance of representative datasets,
comparing them with random datasets.
%and comparing it with a random dataset
%to highlight the benefits of ensuring the accuracy loss before training.}
%pointing out the fact that  it is similar to train a perceptron with a dataset or with its  representative dataset.
In Section \ref{sec:experiments}, we illustrate experimentally the same fact for several multi-layer neural networks.
Finally, some conclusions and future work are provided in Section \ref{sec:conclusions}. 

\section{Background}\label{sec:background}
Next we recall some basic definitions and notations used throughout the paper.

\subsection{Neural Networks}
The research field of neural networks is extremely vivid and new architectures are continuously being presented (see, e.g., {\it CapsNets} \cite{1512.03385}, \ma{{\it bidirectional feature pyramid networks} \cite{DBLP:journals/corr/abs-1911-09070}} or new variants of the {\it Gated Recurrent Units} \cite{1701.05923,1701.03452}), so the current notion of neural network is far from the classic multilayer perceptron or radial basis function networks \cite{haykin2009neural}. 

As a general setting, a neural network is a mapping ${\cal N}_{w,\Phi}: \mathbb{R}^n \to \mathbb{R}^m$ that depends on a set of weights $w$ and a set of parameters $\Phi$ which involves the description of the synapses between neurons, layers, activation functions and whatever consideration in its architecture. 
To train the neural network ${\cal N}_{w,\Phi}$,
we use a {\it dataset} 
which is a finite set of pairs 
${\cal D}= \{(x,c_x) \text{ where, 
point} x \text{ lies in } X\subset \mathbb{R}^n \text{ and label } c_x \text{ lies in } \{0,1,\dots,k\}\}$, for a fixed integer $k\in\N$.
The sets $X$ and $\{0,1,\dots,n\}$ are called, respectively, the set of points and the set of labels in ${\cal D}$.

To perform the learning process, we use: (1) a {\it loss function} which measures the difference between the output of the network (obtained with the current weights) and the desired output; and (2) a loss-driven {\it training} method to iteratively update the weights.

\subsection{Persistent Homology}\label{sec:perHom}

In this paper, the representativeness of a dataset will be measured using methods from the recent developed area called Computational Topology whose main tool is {\it persistent homology}.
A detailed presentation of this field can be found in \cite{Edelsbrunner10}.

Homology provides mathematical formalism to study how a space is connected, being the $q$-dimensional homology group the mathematical representation for the $q$-dimensional ``holes'' in a given space. This way, the 0-dimensional  homology group counts the connected components of the space, the 1-dimensional homology group its tunnels and the 2-dimensional homology group its  cavities. For higher dimensions, the intuition of holes is lost. 

Persistent homology is usually computed when the homology groups can not be determined. An example of the latter appear when a surface is sampled by a point cloud.
Persistent homology is based on the concept of {\it filtration}, which is an increasing sequence of {\it simplicial complexes} which ``change'' along time. 
The building blocks of a simplicial complex are $q$-simplices.

\begin{definition}
The $q$-simplex generated by $q+1$  affinely independent points $v^0,\dots, v^q\in \mathbb{R}^n$  (being $q\leq n$),    denoted by  $\mu = \left< v^0,\dots,v^q\right>$,   is defined by:
  $$\mu=\left\{x\in \mathbb{R}^n\; \mid \; x =\sum_{i=0}^q \lambda_i {v^i}:\;  
  \lambda_i \ge 0,\, \sum_{i=0}^q \lambda_i = 1\right\}.. $$
 A   face $\tau$ of $\mu$ is a simplex  generated by any subset of $\{v^0,\dots,v^q\}$.  In such case we denote $\tau \le \mu$.
\end{definition}

From the notions of $q$-simplex and face, the concepts of simplicial complex, subcomplex and filtration arise in a natural way.

\begin{definition}
  A simplicial complex $K$ is a finite set of simplices, satisfying
  the following properties:
  \begin{enumerate}
  \item If $\mu, \tilde{\mu} \in K$ and $\mu \cap \tilde{\mu}
    \neq \emptyset$, then $\mu \cap \tilde{\mu} \in K$.
  \item If $\mu \in K$ and $\tau \le \mu$, then $\tau \in K$. 
  \end{enumerate}

A subset $\tilde{K}$  of $K$ is 
a  subcomplex of   $K$ if $\tilde{K}$ is also a simplicial complex.

 A filtration of a simplicial complex $K$ is a nested sequence of
subcomplexes,
 $$\emptyset = K_{0} \subset K_{1} \subset \dots \subset K_m = K. $$
\end{definition}

An example of filtration is the Vietoris-Rips  filtration (see \cite{Hausmann1994}). Roughly speaking, a Vietoris-Rips filtration is obtained by ``growing'' open balls centered at every point of a given set in an $n$-dimensional space. 
Specifically, the filtration is built by increasing with time the radius of the balls and joining those vertices whose balls intersect forming new simplices. 

This way, we say that a $q$-dimensional hole {\it is born} along the filtration
when it  appears and  we say that a $q$-dimensional hole {\it dies} when it
 merges with another $q$-dimensional hole at a certain time during the construction of the  filtration.
 The rule to decide which hole dies when they merge is the {\it elder rule} which establishes that younger holes die.
So, births and deaths of all the holes
are controlled over time in the filtration. One of the common graphical representations of births and deaths of the $q$-dimensional holes over time
is the so-called ($q$-dimensional) {\it persistence diagram} which consists of a set of points on the Cartesian plane. A point of a persistence diagram represents the birth and the death of a $q$-dimensional hole. 
Since deaths happen only after births, all the points in  a persistence diagram lie above the diagonal axis. Furthermore, those points in  a persistence diagram that are far from the diagonal axis are  candidates to be ``topologically significant'' since they represent holes that survive for a long time. The so-called bottleneck distance can be used to compare two persistence diagrams.

\begin{definition}
The ($q$-dimensional) bottleneck distance between two ($q$-dimensional) persistence diagrams $\Dgm$ and $\widetilde{\Dgm}$ is:
$$d_B(\Dgm,\widetilde{\Dgm})= \inf_{\phi } \;\sup_{\alpha} \| \alpha-\phi(\alpha) \|_{\infty} $$
where $\alpha\in \Dgm$ and
$\phi$ is any possible bijection between $\Dgm\cup\Delta$ and $\widetilde{\Dgm}\cup\Delta$, being $\Delta$ the set of points in the diagonal axis.
\end{definition}

An useful result used in this paper is the following one that connects  the Gromov-Hausdorff distance between two metric spaces and  the bottleneck distance between the persistence diagrams obtained from their corresponding Vietoris-Rips filtrations. 
For the sake of brevity, the ($q$-dimensional) persistence diagram obtained from the Vietoris-Rips filtration computed from a subset $X$ of $\mathbb{R}^n$, with $q\leq n$, will be  simply called  the ($q$-dimensional) persistence diagram of $X$ and denoted by $\Dgm_q(X)$.

\begin{theorem}\label{th:stab}\cite[Theorem 5.2]{Chazal2014}
For any two subsets $X$ and $Y$ of $\mathbb{R}^n$, and for any dimension $q\leq n$, the bottleneck distance between
the persistence diagrams  of $X$ and $Y$, $\Dgm_q(X)$ and $\Dgm_q(Y)$, is bounded by the Gromov-Hausdorff distance of $X$ and $Y$:
$$d_B(\Dgm_q(X),\Dgm_q(Y)) \le 2d_{GH}(X,Y).$$
\end{theorem}

Let us recall that the Hausdorff distance between 
$X$ and $Y$ is:
$$d_H (X,Y) = \max \{ \,\sup_{x} \;\inf_{y} \;\|x-y\| ,\sup_{y} \;\inf_{x}\; \|x-y\|\, \}, $$
where $x\in X$ and $y\in Y$, and the Gromov-Hausdorff distance between $X$ and $Y$ is defined as the infimum of the Hausdorff distance taken over all possible isometric transformations \cite{Chazal2014}. That is,
$$d_{GH}(X,Y) =\frac{1}{2} \inf_{f,g}\{\,d_H (f(X), g(Y))\,   \}, $$
where $f:X\rightarrow Z$ denotes an isometric transformation of $X$ into some metric space $Z$ and  $g:Y\rightarrow Z$ denotes an isometric transformation of $Y$ into $Z$. 

\section{Representative Datasets}\label{sec:repdatasets}
As mentioned above, the key idea in this paper is the formal definition of representative datasets and the proof of their usefulness to reduce the 
resources needed to train a neural network without 
losing accuracy. Proving such a general result for  any neural network architecture is out of the scope of this paper and, probably, it is not possible since the definition of neural network is continuously evolving over time as mentioned before. 
Due to such difficulties, we will begin in this paper by proving the usefulness of representative datasets in the perceptron case, and by experimentally showing their effectiveness in the case of multilayer neural networks.

To start with, in this section, we provide the definition of representative datasets which is independent of the neural network architecture considered.
The intuition behind this definition is to keep the ``shape'' of the original dataset while reducing its number of points.
Firstly, let us introduce the notion of $\varepsilon$-representative point.

\begin{definition}
A labeled point $(\tilde{x},c_{\tilde{x}})\in \mathbb{R}^n\times \{0,1,\dots,k\}$ is $\varepsilon$-representative of $(x,c_x)\in \mathbb{R}^n\times \{0,1,\dots,k\}$ if 
$c_x=c_{\tilde{x}}$ and there exists $\delta \in \mathbb{R}^n$ such that $x=\tilde{x}+\delta$ with $\|\delta\|\le \varepsilon$, where $\varepsilon\in \mathbb{R}$ is the representation error. We denote $\tilde{x}\approx_\varepsilon x$.
\end{definition}

The next step is to define the concept of $\varepsilon$-representative dataset.
Notice that if a dataset can be correctly classified by a neural network, any isometric transformation of such dataset can be
also correctly classified by the neural network (after adjusting the weights).
Therefore, the definition of $\varepsilon$-representative dataset should be independent of such transformations.
Moreover, the concept of $\lambda$-balanced
$\varepsilon$-representative datasets is also introduced and it will be used in Section \ref{sec:perceptroncase} to ensure that similar results \ma{are obtained} when training a perceptron  with a representative dataset instead of training it with the original \ma{one.}

\begin{definition}
A dataset $\tilde{{\cal D}}= \{(\tilde{x},c_{\tilde{x}}) : \tilde{x}\in \tilde{X}\subset \mathbb{R}^n$ and $c_{\tilde{x}}\in \{0,1,\dots,k\}\}$  
is  $\varepsilon$-representative of ${\cal D} = \{(x,c_x): x\in X\subset \mathbb{R}^n
$ and $c_x\in \{0,1,\dots,k\}\}$ if there exists an isometric transformation $f:\tilde{X}\to \mathbb{R}^n$, 
such that for any $(x,c_x)\in {\cal D}$ there exists $(\tilde{x},c_{\tilde{x}})\in \tilde{{\cal D}}$
satisfying that  $f(\tilde{x})\approx_\varepsilon x$.
The  dataset $\tilde{{\cal D}}$ is said to be  $\lambda$-balanced      
if for 
each $(\tilde{x},c_{\tilde{x}})\in {\cal \tilde{D}}$, the set $
\{(x,c_x) : f(x)\approx_\varepsilon\tilde{x} \}
$
contains  $\lambda$ points.
Finally, we will say that $\varepsilon$ is {\it optimal} if 
$\varepsilon=\min\{\xi:$
there exists a $\xi$-representative dataset of ${\cal D}\}$.
\end{definition}

\begin{proposition}\label{prop:gh}
Let $\tilde{{\cal D}}$ be an  $\varepsilon$-representative dataset (with set of points $\tilde{X}$)
of  a dataset ${\cal D}$ (with set of points $X$).
Then $d_{GH}(X,\tilde{X})\le \varepsilon$.
\end{proposition}
\begin{proof}
By definition of $\varepsilon$-representative datasets, there exists an isometric transformation from $\tilde{X}$ to 
$\mathbb{R}^n$ where for all $x\in X$ there exists $\tilde{x}\in\tilde{X}$ such that $\|x-f(\tilde{x})\| \le \varepsilon$. Therefore, $d_H(X,f(\tilde{X}))\le \varepsilon$. Then, by  the definition of the Gromov-Hausdorff distance, 
$$d_{GH}(X,\tilde{X})\le d_H(X,f(\tilde{X}))\le \varepsilon.$$
\end{proof}

The definition of $\varepsilon$-representative datasets is not useful when $\varepsilon$ is ``big''. The following result, which is a consequence of Proposition~\ref{prop:gh},
provides the optimal value for $\varepsilon$.

\begin{corollary}
If $\varepsilon=d_{GH}(X,\tilde{X})$ then $\varepsilon$ is optimal.
\end{corollary}

Therefore, one way to discern if a dataset $\tilde{{\cal D}}$ is ``representative enough'' of ${\cal D}$ is to compute the Gromov-Hausdorff distance between $X$ and $\tilde{X}$. If the Gromov-Hausdorff distance is ``big'', we could say that the dataset $\tilde{{\cal D}}$ is not representative of ${\cal D}$. 
However, the Gromov-Hausdorff distance is not useful in practice because of its high computational cost. An alternative approach to this problem is given in Section \ref{section:TDA}. 

Finally, the definition of {\it dominating set} is introduced, which will be used in the next subsection.

\begin{definition}
Given a graph $G=(X,E)$, a set $Y\subset X$ is a dominating set of $X$ if for any $x\in X$, it is satisfied that $x\in Y$ or there exists $y\in Y$  adjacent to $x$.
\end{definition}

\subsection{Proximity Graph Algorithm}
In this subsection, for a given $\varepsilon>0$, we propose a variant of the Proximity Graph Algorithm \cite{ENDM2737} 
to compute an  $\varepsilon$-representative dataset  $\tilde{\cal D}$  
of a  dataset ${\cal D}=\{(x,c_x):x\in X 
\mbox{ and } c_x\in \{0,1,\dots,k\} \}$.

Firstly, a proximity graph is built over $X$, establishing  adjacency relations between the points of $X$, represented by arcs.

\begin{figure}[t]
\centering
        \includegraphics[width = 0.45 \textwidth]{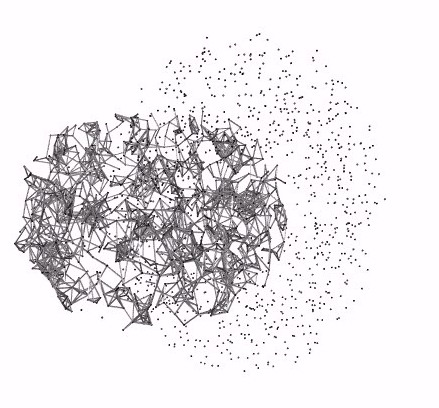}
        \caption{A point cloud sampling two interlaced solid torus and the $\varepsilon$-proximity graph of one of them for a fixed $\varepsilon$.}
        \label{fig:pred_surface}
\end{figure}

\begin{definition}
Given $X\subset \mathbb{R}^n$ and $\varepsilon>0$, 
an $\varepsilon$-proximity graph of $X$ is a graph $G_\varepsilon(X)=(X,E)$ such that if $x,y\in X$ and $\|x-y\|\le \varepsilon$ then $(x,y)\in E$.
\end{definition}
See Fig.~\ref{fig:pred_surface} in which
the proximity graph of one of the 
two interlaced solid torus is drawn for a fixed $\varepsilon$.

\ma{Secondly},  from a $\varepsilon$-proximity graph of $X$, a dominating set $\tilde{X}\subseteq X$ is computed, obtaining an $\varepsilon$-representative dataset $\tilde{\cal D} = \{(\tilde{x},c_{\tilde{x}}):\tilde{x}\in \tilde{X}$ and $(\tilde{x},c_{\tilde{x}})\in {\cal D}\}$ also called  {\it dominating dataset} of ${\cal D}$.
Algorithm~\ref{algo} shows the  pseudo-code used in this paper to compute the dominating dataset of ${\cal D}$.

\begin{algorithm}
\caption{Dominating Dataset Algorithm}\label{algo}
\hspace*{\algorithmicindent} \textbf{Input} A dataset ${\cal D} = \{(x,c_x) : x\in  X\subset \mathbb{R}^n \mbox{ and }c_x\in \{0,1,\dots,k\}\}$ and a parameter $\varepsilon>0$ \\
\hspace*{\algorithmicindent} \textbf{Output} A dataset ${\cal \tilde{D}}\subseteq {\cal D}$
\begin{algorithmic}[1]
\For  {$c=0$ \textbf{to} $c=k$} 
\State $X_c = \{x: (x,c)\in {\cal D}\}$
\State $\tilde{X}_c=$ DominatingSet$(G_\varepsilon(X_c))$
\EndFor
\State   $\tilde{X} = \cup_{c=0}^k \tilde{X}_c$ 
\State   $\tilde{\cal D} = \big\{(x,c): x\in \tilde{X} \mbox{ and } (x,c)\in {\cal D}\big\}$
\end{algorithmic}
\end{algorithm}
Here, DominatingSet$(G_\varepsilon(X_c))$ refers to a dominating set obtained from the proximity graph $G_\varepsilon(X_c)$. Among the existing 
algorithms in the literature to obtain a dominating set, we will use, in our experiments in Section \ref{sec:experimentalresults}, the algorithm proposed in \cite{Matula87} that runs in $O(\lvert X \rvert\cdot |E|)$.
Therefore, the complexity of  Algorithm \ref{algo} is
$O(\lvert X \rvert^2+\lvert X \rvert\cdot |E|)$ because of the size of the matrix of distances between points and the complexity of the algorithm to obtain the dominating set. 

\begin{lemma}
The dominating dataset ${\cal \tilde{D}}$ 
obtained by running Algorithm \ref{algo} is an $\varepsilon$-representative dataset  of ${\cal D}$.
\end{lemma}

\begin{proof}
Let us prove that for any $(x,c_x)\in {\cal D}$ there exists $(\tilde{x},c_{x})\in \tilde{{\cal D}}$ such that $x \approx_{\varepsilon} \tilde{x}$. Two possibilities  arise:
\begin{enumerate}
    \item If $(x,c_x)\in \tilde{{\cal D}}$, it is done.
    \item If $(x,c_x)\not \in \tilde{{\cal D}}$, 
    since $\tilde{X}_{c_x}\subset \tilde{X}$ is a dominating set of $G_\varepsilon(X_{c_x})$, then 
    there exists $\tilde{x}\in \tilde{X}_{c_x}$ such that $\tilde{x}$ is adjacent to $x$ in $G_\varepsilon(X_{c_x})$. Therefore,  $(\tilde{x},c_x)\in \tilde{{\cal D}}$ and  $x\approx_{\varepsilon}\tilde{x}$.
\end{enumerate}
\end{proof}

\section{
Persistent homology to infer the representativeness of a dataset}\label{section:TDA}

In this section, we recall the role of persistent homology as a tool to infer the representativeness of a dataset.

Firstly, from Theorem~\ref{th:stab} in page \pageref{th:stab}, we can establish that  the bottleneck distance between persistence diagrams is a lower bound of the representativeness of the dataset.

\begin{lemma}
\label{le:bottleneck}
Let $\tilde{\cal D}$ be an  $\varepsilon$-representative dataset (with set of points $\tilde{X}\subset  \mathbb{R}^n$) 
of a dataset  ${\cal D}$ (with set of points $X\subset  \mathbb{R}^n$). Let $\Dgm_q(X)$ and $\Dgm_q(\tilde{X})$ be the $q$-dimensional persistence diagrams of $X$ and $\tilde{X}$, respectively. Then, for $q\leq n$,
$$d_{B}\big(\Dgm_q(X),\Dgm_q(\tilde{X})\big)\le 2 \varepsilon.$$
\end{lemma}

\begin{proof}
Since $\tilde{\cal D}$ is an $\varepsilon$-representative  dataset of ${\cal D}$ then  $d_{GH}(X$, $\tilde{X})\leq \varepsilon$ by 
 Proposition \ref{prop:gh}. Now, by Theorem \ref{th:stab}, $$\frac{1}{2}d_B(\Dgm_q(X),\Dgm_q(Y)) \le d_{GH}(X,Y)\le \varepsilon.$$
\end{proof}

As a direct consequence of Lemma~\ref{le:bottleneck} and the fact that the Hausdorff distance is an upper bound of the Gromov-Hausdorff distance, we have the following.

\begin{corollary}\label{cor:interval}
Let ${\cal \tilde{D}}$ be  an $\varepsilon$-representative dataset 
of ${\cal D}$ where the parameter $\varepsilon$ is optimal. Let $\Dgm_q(X)$ and $\Dgm_q(\tilde{X})$ be
the $q$-dimensional persistence diagrams of $X$ and $\tilde{X}$, respectively.  Then, 
$$\frac{1}{2}d_B\big(\Dgm_q(X),\Dgm_q(\tilde{X}))\big)\leq\varepsilon\leq d_H\big(X,\tilde{X}\big).$$
\end{corollary}

In order to illustrate the usefulness of this last result, we will discuss a simple example. In Fig.~\ref{fig:binary_or}, we can see a  subsample of  a circumference (the original dataset) together with two classes corresponding, respectively,  to the upper and lower part of the circumference.
In Fig.~\ref{binary_not}, we can see a 
subset of the original dataset and a decision boundary ``very''different to the one given in  Fig.~\ref{fig:binary_or}. Then we could say that the dataset showed in Fig.~\ref{binary_not}  does not ``represent'' the same classification problem
than the original dataset. 
However, the dataset shown  in Fig.~\ref{binary_rep} could be considered a  representative dataset of the original one since both  decision boundaries are  ``similar''.
This can be determined by computing 
the Hausdorff distance between the original and the other datasets, and the bottleneck distance between the persistence diagrams of the corresponding datasets  (see the values shown in Table \ref{metrics_example}). 
Using Corollary \ref{cor:interval}, we can infer that $0.08\leq \varepsilon_1\leq 0.18$ for the dataset given in Fig.~\ref{binary_rep} and $0.13\leq \varepsilon_2\leq 0.3$ for the dataset given in Fig.~\ref{binary_not}. Therefore, the dataset given in Fig.~\ref{binary_rep} can be considered ``more'' representative of the dataset showed in  Fig.~\ref{fig:binary_or} than the dataset given in Fig.~\ref{binary_not}, as expected.

\begin{figure}[!h]
\minipage{0.3 \textwidth}
\definecolor{qqqqff}{rgb}{0.,0.,1.}
\definecolor{ffqqqq}{rgb}{1.,0.,0.}
\definecolor{xdxdff}{rgb}{0.49019607843137253,0.49019607843137253,1.}
\definecolor{ududff}{rgb}{0.30196078431372547,0.30196078431372547,1.}
\begin{tikzpicture}[scale = 0.035,line cap=round,line join=round]
\clip(-51.072829243626984,-35.377220802765706) rectangle (64.16507297668906,38.107818294247934);
\draw [line width=2.pt,domain=-51.072829243626984:64.16507297668906] plot(\x,{(-9.478023899026002--0.31314647342477403*\x)/61.79423742248831});
\draw[line width=2.pt,dash pattern=on 1pt off 1pt,color=qqqqff,fill=qqqqff,fill opacity=0.25](-51.072829243626984,38.107818294247934)--(-51.59474003266827,-0.4122719354820896)--(64.68698376573035,0.17111245223424207)--(64.16507297668906,38.107818294247934);
\draw[line width=2.pt,dash pattern=on 1pt off 1pt,color=ffqqqq,fill=ffqqqq,fill opacity=0.25](64.16507297668856,0.16849403038632071)--(-51.07282924362649,-0.4096535136341683)--(-51.072829243626984,-0.4096535136341683)--(-51.072829243626984,-35.377220802765706)--(64.16507297668906,-35.377220802765706)--(64.16507297668906,0.16849403038632071);
\begin{scriptsize}
\draw [fill=xdxdff] (0.,30.) circle (30pt);
\draw [fill=xdxdff] (-10.312493000590392,28.1718385646513) circle (30pt);
\draw [fill=xdxdff] (-19.8453371647693,22.49805753429896) circle (30pt);
\draw [fill=xdxdff] (-25.354845801060154,16.035329569561778) circle (30pt);
\draw [fill=xdxdff] (-28.319022422546297,9.901159984129128) circle (30pt);
\draw [fill=xdxdff] (-29.75989725176625,3.7879434478771694) circle (30pt);
\draw [fill=xdxdff] (9.761370245042563,28.36751048187341) circle (30pt);
\draw [fill=xdxdff] (20.057095075834262,22.309480879638258) circle (30pt);
\draw [fill=xdxdff] (25.95730472236016,15.04055622477237) circle (30pt);
\draw [fill=xdxdff] (28.312687514014545,9.919260342070116) circle (30pt);
\draw [fill=xdxdff] (29.7457006844596,3.89785720499607) circle (30pt);
\draw [fill=ffqqqq] (-28.326269003223416,-9.88040911890921) circle (30pt);
\draw [fill=ffqqqq] (-22.31161658929157,-20.05471927433663) circle (30pt);
\draw [fill=ffqqqq] (-10.146321811993474,-28.232112101071458) circle (30pt);
\draw [fill=ffqqqq] (0.,-30.) circle (30pt);
\draw [fill=ffqqqq] (-29.704206506242432,-4.202394059878807) circle (30pt);
\draw [fill=ffqqqq] (-25.311134705144372,-16.104237328667136) circle (30pt);
\draw [fill=ffqqqq] (9.912333635871622,-28.31511331234908) circle (30pt);
\draw [fill=ffqqqq] (19.957322677713066,-22.39877834922342) circle (30pt);
\draw [fill=ffqqqq] (25.9972777605403,-14.97135762184969) circle (30pt);
\draw [fill=ffqqqq] (28.114656436010485,-10.467382360700036) circle (30pt);
\draw [fill=ffqqqq] (29.780169369759083,-3.625122385308261) circle (30pt);
\end{scriptsize}
\end{tikzpicture}
\subcaption{A binary classification problem given by a sampled circumference. In this case, the classification problem tries to distinguish between the upper  and the lower part of the circumference.}
\label{fig:binary_or}
\endminipage \hfill
\minipage{0.3 \textwidth}
\definecolor{qqqqff}{rgb}{0.,0.,1.}
\definecolor{ffqqqq}{rgb}{1.,0.,0.}
\definecolor{xdxdff}{rgb}{0.49019607843137253,0.49019607843137253,1.}
\definecolor{ududff}{rgb}{0.30196078431372547,0.30196078431372547,1.}

\begin{tikzpicture}[scale = 0.035,line cap=round,line join=round]
\clip(-51.072829243626984,-35.377220802765706) rectangle (64.16507297668906,38.107818294247934);
\draw [line width=2.pt,domain=-51.072829243626984:64.16507297668906] plot(\x,{(--82.9691590704375--3.2358468920559984*\x)/64.09064489426997});
\draw[line width=2.pt,dash pattern=on 1pt off 1pt,color=qqqqff,fill=qqqqff,fill opacity=0.25](-51.072829243626984,38.107818294247934)--(-51.59474003266827,-1.313730750667671)--(64.68698376573035,4.564764149017568)--(64.16507297668906,38.107818294247934);
\draw[line width=2.pt,dash pattern=on 1pt off 1pt,color=ffqqqq,fill=ffqqqq,fill opacity=0.25](64.16507297668907,4.538379521999234)--(-51.072829243627005,-1.2873461236493362)--(-51.072829243626984,-1.2873461236493362)--(-51.072829243626984,-35.377220802765706)--(64.16507297668906,-35.377220802765706)--(64.16507297668906,4.538379521999234);
\begin{scriptsize}
\draw [fill=xdxdff] (0.,30.) circle (30pt);
\draw [fill=xdxdff] (-19.8453371647693,22.49805753429896) circle (30pt);
\draw [fill=xdxdff] (-29.75989725176625,3.7879434478771694) circle (30pt);
\draw [fill=xdxdff] (20.057095075834262,22.309480879638258) circle (30pt);
\draw [fill=xdxdff] (28.312687514014545,9.919260342070116) circle (30pt);
\draw [fill=ffqqqq] (-10.146321811993474,-28.232112101071458) circle (30pt);
\draw [fill=ffqqqq] (-29.704206506242432,-4.202394059878807) circle (30pt);
\draw [fill=ffqqqq] (9.912333635871622,-28.31511331234908) circle (30pt);
\draw [fill=ffqqqq] (25.9972777605403,-14.97135762184969) circle (30pt);
\draw [fill=ffqqqq] (29.780169369759083,-3.625122385308261) circle (30pt);
\end{scriptsize}
\end{tikzpicture}
\subcaption{{\bf ($\varepsilon_1$-Representative dataset)} A subset of the sampled circumference given in Fig. \ref{fig:binary_or}. 
Let us observe that the decision boundary obtained is similar to the one showed in Fig. \ref{fig:binary_or}.}
\label{binary_rep}
\endminipage \hfill
\minipage{0.3 \textwidth}
\definecolor{qqqqff}{rgb}{0.,0.,1.}
\definecolor{ffqqqq}{rgb}{1.,0.,0.}
\definecolor{xdxdff}{rgb}{0.49019607843137253,0.49019607843137253,1.}
\begin{tikzpicture}[scale = 0.035,line cap=round,line join=round]
\clip(-51.072829243626984,-35.377220802765706) rectangle (64.16507297668906,38.107818294247934);
\draw [line width=2.pt,domain=-51.072829243626984:64.16507297668906] plot(\x,{(--32.29198491672133-31.056631426026165*\x)/35.85302237985239});
\draw[line width=2.pt,dash pattern=on 1pt off 1pt,color=qqqqff,fill=qqqqff,fill opacity=0.25](-42.95894680387428,38.107818294247934)--(41.886095628685055,-35.377220802765706)--(64.16507297668906,-35.377220802765706)--(64.16507297668906,38.107818294247934);
\draw[line width=2.pt,dash pattern=on 1pt off 1pt,color=ffqqqq,fill=ffqqqq,fill opacity=0.25](-43.547513719429986,38.107818294247934)--(-51.072829243626984,38.107818294247934)--(-51.072829243626984,-35.377220802765706)--(42.475009762722244,-35.377220802765706)--(-43.547513719429986,38.107818294247934);
\begin{scriptsize}
\draw [fill=xdxdff] (9.761370245042563,28.36751048187341) circle (30pt);
\draw [fill=xdxdff] (20.057095075834262,22.309480879638258) circle (30pt);
\draw [fill=xdxdff] (25.95730472236016,15.04055622477237) circle (30pt);
\draw [fill=xdxdff] (28.312687514014545,9.919260342070116) circle (30pt);
\draw [fill=xdxdff] (29.7457006844596,3.89785720499607) circle (30pt);
\draw [fill=ffqqqq] (-28.326269003223416,-9.88040911890921) circle (30pt);
\draw [fill=ffqqqq] (-22.31161658929157,-20.05471927433663) circle (30pt);
\draw [fill=ffqqqq] (-10.146321811993474,-28.232112101071458) circle (30pt);
\draw [fill=ffqqqq] (-29.704206506242432,-4.202394059878807) circle (30pt);
\draw [fill=ffqqqq] (-25.311134705144372,-16.104237328667136) circle (30pt);
\end{scriptsize}
\end{tikzpicture}
\subcaption{{\bf ($\varepsilon_2$-Representative dataset)} A subset of the sampled circumference given in Fig. \ref{fig:binary_or}. 
Let us observe that the decision boundary obtained is quite different to the one showed in Fig. \ref{fig:binary_or}.
}
\label{binary_not}
\endminipage 

\caption{Illustration of a binary classification problem and the representative dataset concept.}
\end{figure}

\begin{table}[!h]
\renewcommand{\arraystretch}{2}
\centering
\begin{tabular}{cccc}
\hline
Datasets                    & $\frac{1}{2}d_{B0}$
%$b_0$ 
& 
%$b_1$ 
$\frac{1}{2}d_{B1}$
& $d_H$
\\ \hline
Original  (Fig.~\ref{fig:binary_or}) and $\varepsilon_1$-representative  (Fig.~\ref{binary_rep})  & 
%0.15
0.07       &      
%0.16 
0.08 & 0.18 \\ \hline
Original  (Fig.~\ref{fig:binary_or}) and $\varepsilon_2$-representative (Fig.~\ref{binary_not}) &   
%0.26
0.13& 
%0.16 
0.08& 0.3               \\ \hline
\end{tabular}
\caption{The 0-dimensional bottleneck distance ($d_{B0}$), the  1-dimensional bottleneck distance ($d_{B1}$), and the Hausdorff distance ($d_H$) between
(the persistence diagrams of) the dataset given in Fig.~\ref{fig:binary_or} and
%, respectively, 
the datasets given in Fig.~\ref{binary_rep} and Fig.~\ref{binary_not}.}
\label{metrics_example}
\end{table}

\section{The Perceptron Case}\label{sec:perceptroncase}

One of the simplest neural network architecture is the perceptron. 
Our goal in this section is to formally prove that training a perceptron with a representative dataset is equivalent, in terms of accuracy, 
to training it  with the original dataset. Specifically, we provide  results using {\it gradient descent} algorithms. 
Notice that in the case of the stochastic gradient descent algorithm, the number of updates needed to reach convergence usually increases with the size of data (see  \cite[Section 5.9]{Goodfellow-et-al-2016}).

For the sake of simplicity, we will restrict our interest to a binary classification problem,
{although our approach is valid for any classification problem.} Therefore, our input is
a {\it binary} dataset ${\cal D} = \big\{(x,c_x):
x\in X\subset \mathbb{R}^n$ and $c_x\in
\{0,1\}\big\}$.

\begin{definition}
A {\it perceptron} ${\cal N}_w:\mathbb{R}^n\to \{0,1\}$, with weights $w     =(w_0,w_1,\dots,w_n)\in \mathbb{R}^{n+1}$,
is defined as:

$$ {\cal N}_w(x) = \left\{ 
        \begin{tabular}{cc}
        	1 & if $y_w(x) \ge \frac{1}{2}$, \\
        	$0$ & otherwise; \\
        \end{tabular}
\right. $$
being $y_{w}:  \mathbb{R}^{n}    \rightarrow \, (0,1)$ defined as
$$y_{w}(x)=\sigma(wx)$$
 where, for $x=(x_1,\dots,x_n) \in \mathbb{R}^n$,  
$$wx=w_0+ w_1x_1+\dots + w_n x_n, $$
and
$\sigma: \mathbb{R} \to (0,1)$, defined as 
$$\displaystyle \sigma(z) =\frac{1}{1+ e^{-z}}, $$
is the {\it sigmoid function}.
\end{definition}

A useful property of the sigmoid function is the easy expression of its derivative. 
Let $\sigma^m$ denote the composition $\sigma^m=\sigma \stackrel{\mbox{\tiny $m$-times}}\cdots \sigma$.

\begin{lemma}\label{lemmaderv}
If $m\in \N$ and $z\in\mathbb{R}$ then $$0<(\sigma^m)'(z)=
m\sigma^{m}(z)(1-\sigma(z))\leq 
\left(\frac{m}{m+1}\right)^{m+1}.$$
\end{lemma}

\begin{proof}
Firstly, let us observe that $(\sigma^m)'(z)=
     m\sigma^{m}(z)(1-\sigma(z))>0$ since 
    $ 0<\sigma(z)<1$ for all $z\in \mathbb{R}$.
Secondly, let us find the local extrema  of $(\sigma^m)'$ by computing the roots of its derivative: 
$$     (\sigma^m)''(z) 
          =m\sigma^{m}(z)(1-\sigma(z))(m-(m+1)\sigma(z)).
$$
Now, $(\sigma^m)''(z)=0$ if and only if $m-(m+1)\sigma(z)=0$.
The last expression vanishes at $z=\log(m)$.
Besides, $(\sigma^m)''(z)>0$ if and only if $m-(m+1)\sigma(z)>0$ which is true for all $z\in (-\infty,\log(m))$. Analogously, $(\sigma^m)''(z)<0$ for all
$z\in (\log(m),+\infty)$, concluding that
$z= \log(m)$ is a global maximum.
Finally, $(\sigma^m)'(\log(m))=\left(\frac{m}{m+1}\right)^{m+1}$.
\end{proof}

In the following lemma, we prove that
the difference between the outputs of the function $y^m_w$ evaluated at a point $x$ and at its $\varepsilon$-representative point $\tilde{x}$ depends on the weights $w$ and the parameter $\varepsilon$.

\begin{lemma}\label{lemma1}
Let $w\in \mathbb{R}^{n+1}$ and $x$, $\tilde{x}\in \mathbb{R}^n$ with 
$\tilde{x}\approx_{\varepsilon} x$. Then, 
$$\|y^m_{w}(\tilde{x})-y^m_{w}(x)\|\le 
\rho_{m} \|w\|_*\varepsilon,$$
where
$$
\rho_{m} =\rho_{(wx,w\tilde{x},m)} = 
\begin{cases}
(\sigma^m)'(z), & \mbox{if } \log(m) < z,\\
(\sigma^m)'(\tilde{z}), & \mbox{if } \tilde{z} < \log(m),\\
(\sigma^m)'(\log(m)), & \mbox{otherwise.}
\end{cases}
$$
with $z= \min\{wx,w\tilde{x}\}$ and $\tilde{z}=\max\{wx,w\tilde{x}\}$.
\end{lemma}

\begin{proof}
Let us assume, without loss of generality, that $wx\leq w\tilde{x}$. Then, using the Mean Value Theorem, there exists $\beta\in (wx,w\tilde{x})$ such that
\begin{equation*}
y^m_{w}(\tilde{x})-y^m_{w}(x) = (\sigma^m)'(\beta)(w\tilde{x}-wx).
\end{equation*}
By Lemma~\ref{lemmaderv}, the maximum of  $(\sigma^m)'$ in the interval $[z,\tilde{z}]$  is reached at   $\log(m)$ if $z<\log(m)<\tilde{z}$, at $z$ if $\log(m)\leq z$, and at 
  $\tilde{z}$ if $\tilde{z}\leq \log(m)$, with $z= \min\{wx,w\tilde{x}\}$ and $\tilde{z}=\max\{wx,w\tilde{x}\}$.
  Consequently, 
  \begin{equation}\label{eq:eq}
\|y^m_{w}(x)-y^m_{w}(x)\|\leq \rho_m \|w(\tilde{x}-x)\|.
\end{equation}
Applying now the H\"older inequality we obtain:
$$  \|w(\tilde{x}-x)\|
    \leq  \|w\|_*\|\tilde{x}-x\|
    \leq \|w\|_*\varepsilon.$$
 Replacing  $\|w(\tilde{x}-x)\|$  by $\|w\|_*\varepsilon$ in Eq. (\ref{eq:eq}), we obtain the desired result.
\end{proof}

The following result is a direct consequence of Lemma \ref{lemmaderv} and Lemma \ref{lemma1}.

\begin{corollary}\label{cor:lemma1}
Let $w\in \mathbb{R}^{n+1}$ and $x$, $\tilde{x}\in \mathbb{R}^n$ with 
$\tilde{x}\approx_{\varepsilon} x$. Then, 
$$\|y^m_{w}(\tilde{x})-y^m_{w}(x)\|\leq 
\left(\frac{m}{m+1}\right)^{m+1}\|w\|_*\varepsilon.$$
\end{corollary}

As previously  pointed out, we will use the stochastic gradient descent algorithm to train the perceptron, trying to minimize the following error function:
$$\mathbb{E}(w,X) = \frac{2}{\lvert X \rvert} \sum_{x\in X}E_x(w),$$
where, for $(x,c_x)\in {\cal D}$ and $w\in\mathbb{R}^{n+1}$,
$$E_x(w) = \frac{1}{2}(c_x - y_{w}(x))^2$$
is the loss function considered in this paper.

Following the stochastic gradient descent algorithm, each iteration  takes  a random point of the dataset,  being able to repeat the same point in different iterations. Let us assume that   the point
 $(x^i,c_i)\in{\cal D}$ is considered in the $i$-th iteration. Then, the weights $w^i=(w^i_0,w^i_1,\dots,w^i_n)$  of the perceptron are   updated according to the following rule:
 \begin{equation}\label{w}
w^{i+1}_j \leftarrow w^i_j- \eta_i \frac{\partial E_{x^i}(w^i)}{\partial w^i_j}, \end{equation}
  where   $\eta_i>0$ is the {\it learning rate}. 
  Let us now notice that 
\begin{equation}\label{derE}
\frac{\partial E_{x}(w)}{\partial w_j}=(c_x-y_w (x))y'_w(x)x_j,
\end{equation}
for $j\in \{1,\dots,n\}$ 
and $x=(x_1,\dots,x_n)$. Besides,
$$\frac{\partial E_{x}(w)}{\partial w_0}=(c_x-y_w (x))y'_w(x).$$
By abuse of notation, the point $x$ will be considered as a point $(x_0,x_1,\dots,x_n)\in \mathbb{R}^{n+1}$ being $x_0=1$. This way, Eq. (\ref{derE})  works for all $j\in 
\{0,1,\dots,n\}$.

Now,  using Eq. (\ref{derE}), Rule (\ref{w}) can be written as  follows:
 \begin{equation}\label{rule}
w^{i+1}
\leftarrow w^i- \eta_i (c_i-y_{w^i} (x^i))y_{w^i}(x^i)(1-y_{w^i}(x^i))x^i
\end{equation}
for $(x^i,c_i)\in {\cal D}$.

 To assure the convergence of the training process to a local minimum, it is enough that,
 for each iteration $i$,
 the learning rate $\eta_i$  satisfies the following conditions (see \cite[Section 8.3.1]{Goodfellow-et-al-2016}):
  \begin{equation*}
     \begin{split}
 \sum_{i=1}^{\infty}\eta_i = \infty \qquad\mbox{ and }\qquad
 \sum_{i=1}^{\infty}\eta^2_i <\infty.
     \end{split}
 \end{equation*}
 
The next result establishes under which conditions $\varepsilon$-representative points are classified under the same label as the points they represent.
 
\begin{lemma}\label{lem:clas}
Let ${\cal \tilde{\cal D}}$ be an $\varepsilon$-representative  dataset  
of the binary dataset ${\cal D}$. Let ${\cal N}_w$ be a perceptron with weights $w\in\mathbb{R}^{n+1}$. 
Let $(x,c)\in {\cal D}$ and $(\tilde{x},c)\in \tilde{\cal D}$ with $\tilde{x}\approx_\varepsilon x$.
If $\varepsilon \leq \frac{\|wx\|}{\|w\|}$ 
then ${\cal N}_w(x)={\cal N}_w(\tilde{x})$.
\end{lemma}
\begin{proof}
First, if $wx=0$ then $\varepsilon=0$, therefore $x=\tilde{x}$ and then ${\cal N}_w(x)={\cal N}_w(\tilde{x})$. 
Now, let us suppose that $wx<0$. 
Then, $y_w(x)<\frac{1}{2}$ and ${\cal N}_w(x)=0$ by definition of perceptron. 
Since $\varepsilon < \frac{\|wx\|}{\|w\|}$ then 
$x$ and $\tilde{x}$ belong to the same semispace in which the space is divided by the hiperplane $wx=0$. Therefore,
$w\tilde{x}<0$, then
$y_w(\tilde{x})<\frac{1}{2}$ and finally ${\cal N}_w(\tilde{x})=0$.
Similarly, if $wx>0$, then 
$\cN_w(x)=1=\cN_w(\tilde{x})$, concluding the proof.

\end{proof} 

By Lemma \ref{lem:clas}, we can state  that if $\varepsilon$ is ``small enough'' 
then the accuracy of
the perceptron ${\cal N}_w$ evaluated on
${\cal D}$ and $\tilde{\cal D}$ will  coincide. 
Let us formally introduce the concept of  {\it accuracy}.
\begin{definition}
The accuracy of the perceptron ${\cal N}_w$ evaluated on the binary dataset ${\cal D}$ is defined as:
 $${\mathbb A}({\cal D},{\cal N}_w)=\frac{1}{\lvert {\cal D}\rvert}\sum_{(x,c_x)\in {\cal D}}I_w(x),$$
where, for any $(x,c_x)\in {\cal D}$, 
$$ {\cal I}_w(x) = \left\{ 
        \begin{tabular}{cc}
        	1 & if $c_x = {\cal N}_w(x)$, \\
        	$0$ & otherwise. \\
        \end{tabular}
\right. $$

\end{definition}
 The following result holds.
\begin{theorem}
 Let ${\cal \tilde{D}}$ be a 
 $\lambda$-balanced $\varepsilon$-representative
 dataset 
 of the binary  dataset ${\cal D}$. Let ${\cal N}_w$ be a perceptron with weights $w\in\mathbb{R}^{n+1}$.
 If $\varepsilon\leq \min\Big\{ \frac{\|wx\|}{\|w\|}: $ $(x,c_x)\in {\cal D}\Big\}$ then 
 $${\mathbb A}({\cal D},{\cal N}_w)={\mathbb A}({\cal \tilde{D}},{\cal N}_w). $$
\end{theorem}

 \begin{proof}
Since $\tilde{\cal D}$ is $\lambda$-balance $\varepsilon$-representative of ${\cal D}$, then $|{\cal D}|=\lambda\cdot \lvert\tilde{\cal D}\rvert$
and we have:
 \begin{equation*}\label{eq:comp_accuracy}
     \begin{split}
     {\mathbb A}({\cal D},{\cal N}_w)-{\mathbb A}({\cal \tilde{D}},{\cal N}_w)&=
      \frac{1}{\lvert {\cal D}\rvert}\sum_{(x,c_x)\in {\cal D}}I_w(x)-\frac{1}{\lvert {\cal \tilde{D}}\rvert}\sum_{(\tilde{x},c_{\tilde{x}})\in {\cal \tilde{D}}}I_w(\tilde{x}) 
         \\
         &=\frac{1}{\lvert {\cal D}\rvert}\sum_{(x,c_x)\in {\cal D}}\left(I_w(x)-\lambda\cdot I_w(\tilde{x})\right)
         \\&
         =\frac{1}{\lvert {\cal D}\rvert}\sum_{(\tilde{x},c_{\tilde x})\in \tilde{\cal D}}\;\sum_{x\approx_\varepsilon\tilde{x}}\left(I_w(x)-I_w(\tilde{x})\right).
     \end{split}
 \end{equation*}
Finally, 
$I_w(x)=I_w(\tilde{x})$ 
for all  $x\approx_{\varepsilon} \tilde{x}$ and $(\tilde{x},c_{\tilde{x}})\in\tilde{\cal D}$
by Lemma \ref{lem:clas}
since  $\varepsilon<\frac{\|wx\|}{\|w\|}$  
for all $(x,c_x)\in {\cal D}$. 

  \end{proof}

Next, let us compare the two errors $\mathbb{E}(w,X)$ and $\mathbb{E}(w,\tilde{X}) $  obtained when considering the binary dataset ${\cal D}$  and  its $\lambda$-balanced $\varepsilon$-representative dataset ${\cal \tilde{D}}$.

\begin{theorem}\label{th:expectation}
Let ${\cal \tilde{D}}$ be a 
 $\lambda$-balanced $\varepsilon$-representative
 dataset 
 of the binary  dataset ${\cal D}$.
Then:
$$\|\mathbb{E}(w,X)-\mathbb{E}(w,\tilde{X})\|
    \le \frac{1}{\lvert X \rvert}\sum_{x\in X} \big( 2c_{x}\rho_1+\rho_2\big)
    \|w (x-\tilde{x})\|
    $$
where $\rho_m$ (being $m=1,2$) was defined in Lemma~\ref{lemma1}, and for each addend, $x\approx_{\varepsilon}\tilde{x}$.
\end{theorem}

\begin{proof}
\begin{equation*}
\begin{split}
    \mathbb{E}(w,X)-\mathbb{E}(w,\tilde{X}) =& \frac{1}{\lvert X\rvert}\sum_{x\in X} (c_x-y_{w}(x))^2
    -\frac{1}{\lvert \tilde{X}\rvert}\sum_{\tilde{x}\in \tilde{X}} (c_{\tilde{x}}-y_{w}(\tilde{x}))^2\\
    =& \frac{1}{\lvert X\rvert\cdot \lvert\tilde{X}\rvert}\Big(\lvert \tilde{X}\rvert\sum_{x\in X}(c_x-y_{w}(x))^2
    -\lvert X\rvert\sum_{\tilde{x}\in \tilde{X}}(c_{\tilde{x}}-y_{w}(\tilde{x}))^2 \Big).
\end{split}   
\end{equation*}
Since ${\cal \tilde{D}}$ is $\lambda$-balanced $\varepsilon$-representative of ${\cal D}$ then  
$\lvert X \rvert=\lambda\, \|\tilde{X}\|$. Then, 
\begin{equation*}
    \begin{split}
       \|\mathbb{E}(w,X)-\mathbb{E}(w,\tilde{X})\|
        &=\frac{1}{\lvert X\rvert}\; \|\sum_{x\in X}  2c_x \big( y_{\tilde{w}}(\tilde{x})-y_{w}(x)\big)
        +y_{w}^2(x)-y_{\tilde{w}}^2(\tilde{x})\big)\|\\
        &\leq 
        \frac{1}{\lvert X\rvert} \sum_{x\in X}  2c_x \| y_{\tilde{w}}(\tilde{x})-y_{w}(x)\|
        +\|y_{w}^2(x)-y_{\tilde{w}}^2(\tilde{x})\|,
    \end{split}
\end{equation*}
where, for each addend, ${\tilde x}\approx_{\varepsilon} x$.
Applying Lemma \ref{lemma1} for $m=1,2$ to the last expression, we get:
$$\|\mathbb{E}(w,X)-\mathbb{E}(w,\tilde{X})\|\leq
\frac{1}{\lvert X\rvert}\sum_{x\in X} \big( 2c_{x}\rho_1+\rho_2\big)
    \|w (x-\tilde{x})\|.$$
\end{proof}

From this last result we can infer the following. We can always fix the parameter $\varepsilon$ ``small enough'' so that the difference between the error obtained when considering the dataset ${\cal D}$ and its $\varepsilon$-representative dataset is ``close'' to zero.

\begin{theorem}
Let $\delta>0$. Let ${\cal \tilde{D}}$ be a 
 $\lambda$-balanced $\varepsilon$-representative
 dataset 
 of the binary  dataset ${\cal D}$. Let ${\cal N}_w$ be a perceptron with weights $w\in\mathbb{R}^{n+1}$.
If $\varepsilon\leq \frac{54}{43\|w\|_*}\delta$, then 
$\|\mathbb{E}(w,X)-\mathbb{E}(w,\tilde{X})\|<\delta$.
\end{theorem}

\begin{proof}
First, $\rho_1\leq \frac{1}{4}$ and $\rho_2\leq \frac{8}{27}$ by Corollary \ref{cor:lemma1}. Second, since $c_x\in\{0,1\}$, then we have:
    \begin{equation*}
    \begin{split}
       \frac{1}{\lvert X \rvert}\sum_{x\in X} \big( 2c_{x}\rho_1+\rho_2\big)
    \|w (x-\tilde{x})\|
        &\leq \frac{1}{\lvert X \rvert} \sum_{x\in X}  \Big( \frac{1}{2}+\frac{8}{27}\Big)\|w (x-\tilde{x})\|\\
        &=\frac{43}{54} \,\frac{1}{\lvert X \rvert} \sum_{x\in X}\|w (x-\tilde{x})\|.
    \end{split}
\end{equation*}
Applying   H\"older inequality to the last expression we get:
$$\frac{1}{\lvert X \rvert}\sum_{x\in X} \big( 2c_{x}\rho_1+\rho_2\big)
    \|w (x-\tilde{x})\|\leq \frac{43}{54}\|w\|_*\varepsilon.$$
    Therefore, by Lemma \ref{th:expectation}, if $\varepsilon\leq \frac{54}{43\|w\|_*}\delta$, then 
$\|\mathbb{E}(w,X)-\mathbb{E}(w,\tilde{X})\|<\delta$ as stated.

    \end{proof}

Summing up, in the case of stochastic gradient descent, we have proved that it is equivalent to train a perceptron with the binary dataset ${\cal D}$ or with its $\lambda$-balanced $\varepsilon$-representative  dataset.
This fact will be highlighted in Section \ref{sec:experimentalresults} for the perceptron case and in Section \ref{sec:experiments} for neural networks with more complex architectures.

\subsection{Experimental Results}\label{sec:experimentalresults}

In this section, two experiments are provided to support our theoretical results for the perceptron case and to illustrate the usefulness of our method. In the first experiment (Subsection~\ref{sec:syn_perceptron}), several synthetic datasets are presented showing different possible casuistics.
In the second experiment (Subsection \ref{subsec:exp2}), the Iris dataset is considered. 

Besides, in the first experiment, random weight initialization is considered and the holdout procedure  is applied (i.e. the datasets were split into training  and test set) to test the generalization capabilities.
In the other experiment, the perceptron is initiated with random weights and trained with three different datasets: the original dataset, a representative dataset
(being the output of  Algorithm \ref{algo})
and a random dataset of the same size as the size of the representative dataset \edu{but the perceptrons are evaluated on the original dataset}. 
These experiments support that a perceptron trained with representative datasets gets similar accuracy to a perceptron trained with the original dataset. Besides, we show that the training time, in the case of the gradient descent training,  is lower when using a representative dataset, and that representative datasets ensure good performance while the random dataset provides no guarantees.
The implementation of the methodology presented here can be consulted online in \url{https://github.com/Cimagroup/Experiments-Representative-datasets}. 

\begin{figure}
\minipage{0.3\textwidth}
\begin{subfigure}{\textwidth}
  \centering
\includegraphics[width=\linewidth]{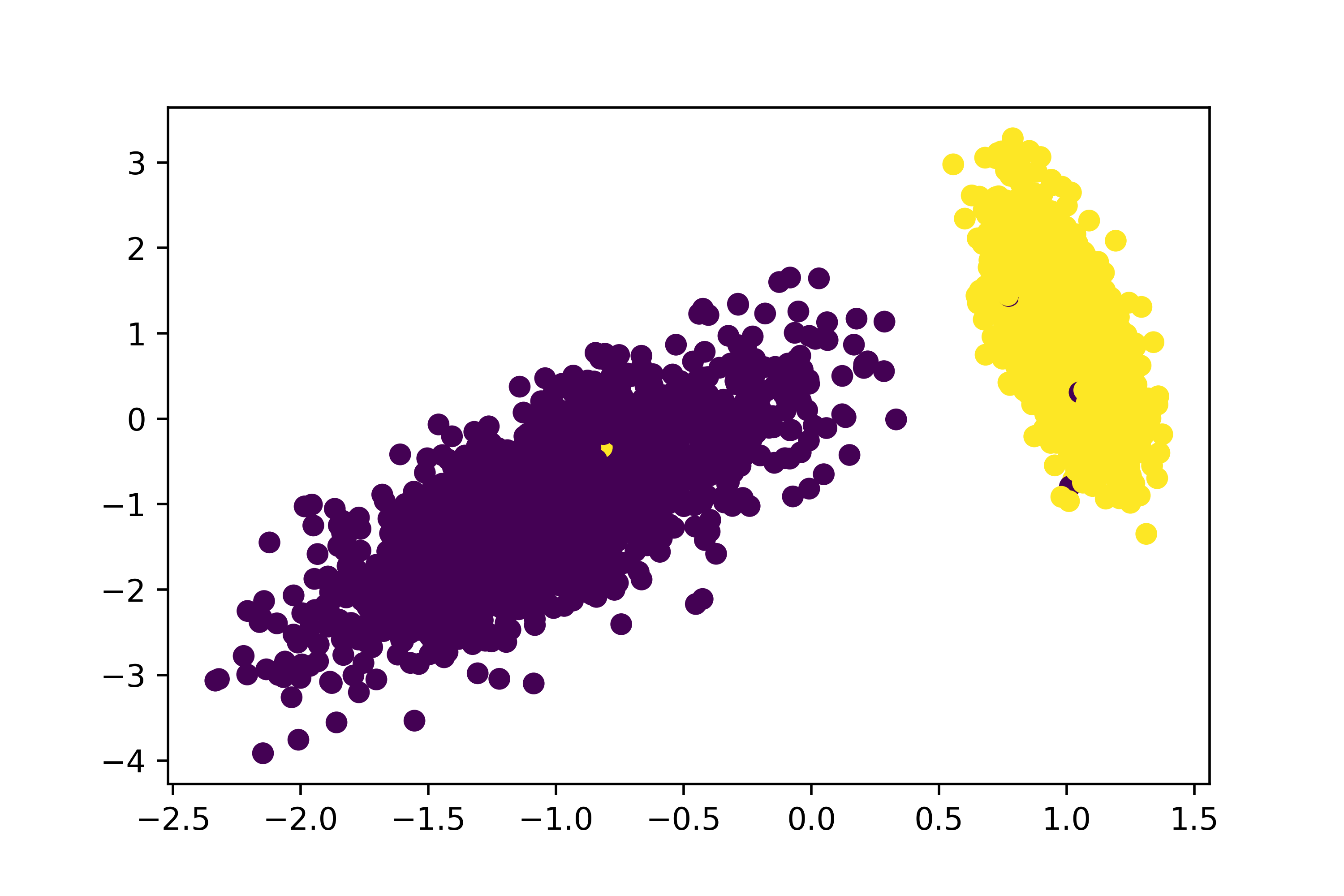}
  \caption{Case 1: original data}
  \label{fig:syn_or_data_1}
\end{subfigure}%
\endminipage\hfill
\minipage{0.3\textwidth}
\begin{subfigure}{\textwidth}
  \centering
  \includegraphics[width=\linewidth]{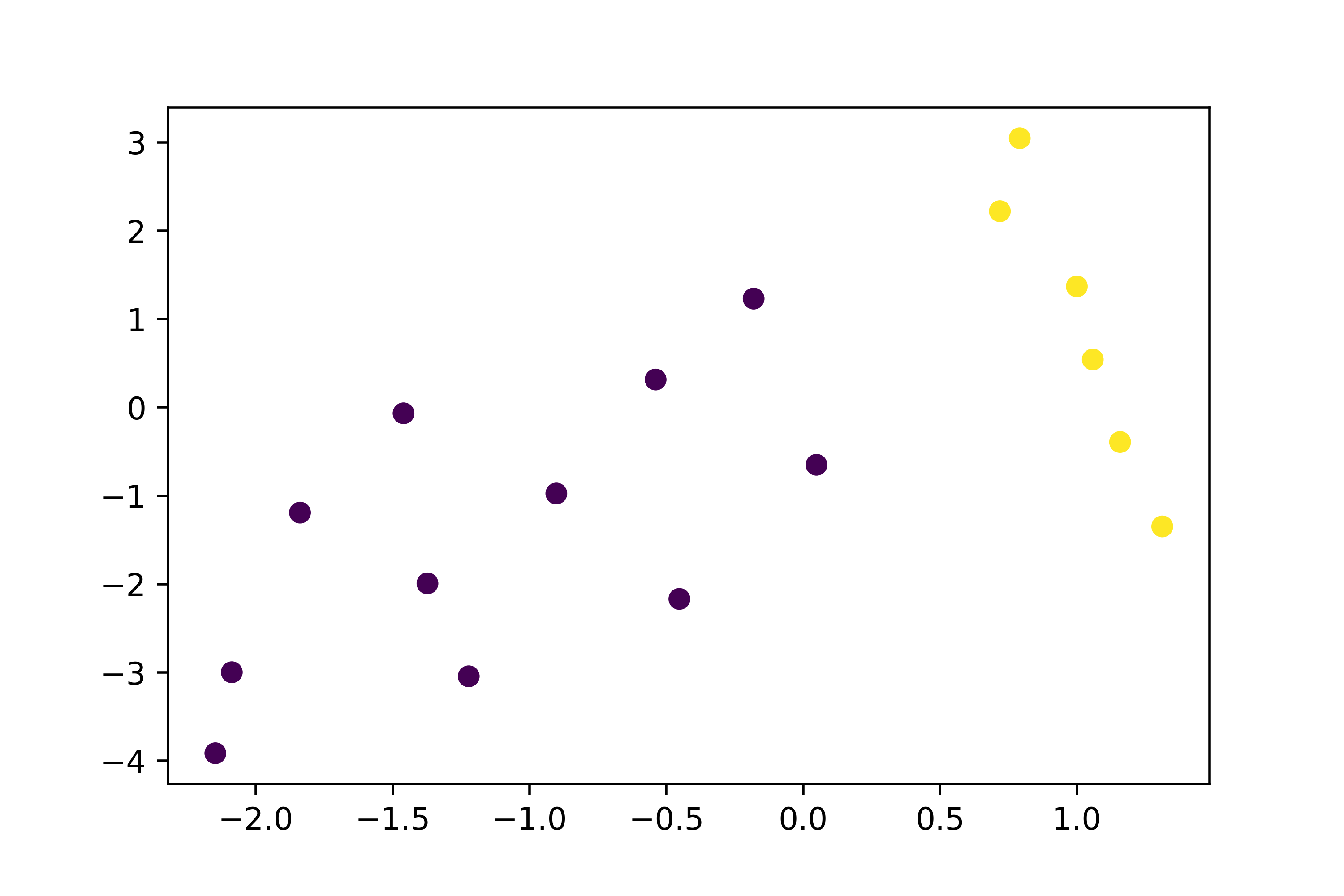}
  \caption{Case 1: dominating 
  dataset}
  \label{fig:syn_dom_data_1}
\end{subfigure}
\endminipage\hfill
\minipage{0.3\textwidth}
\begin{subfigure}{\textwidth}
  \centering
  \includegraphics[width=\linewidth]{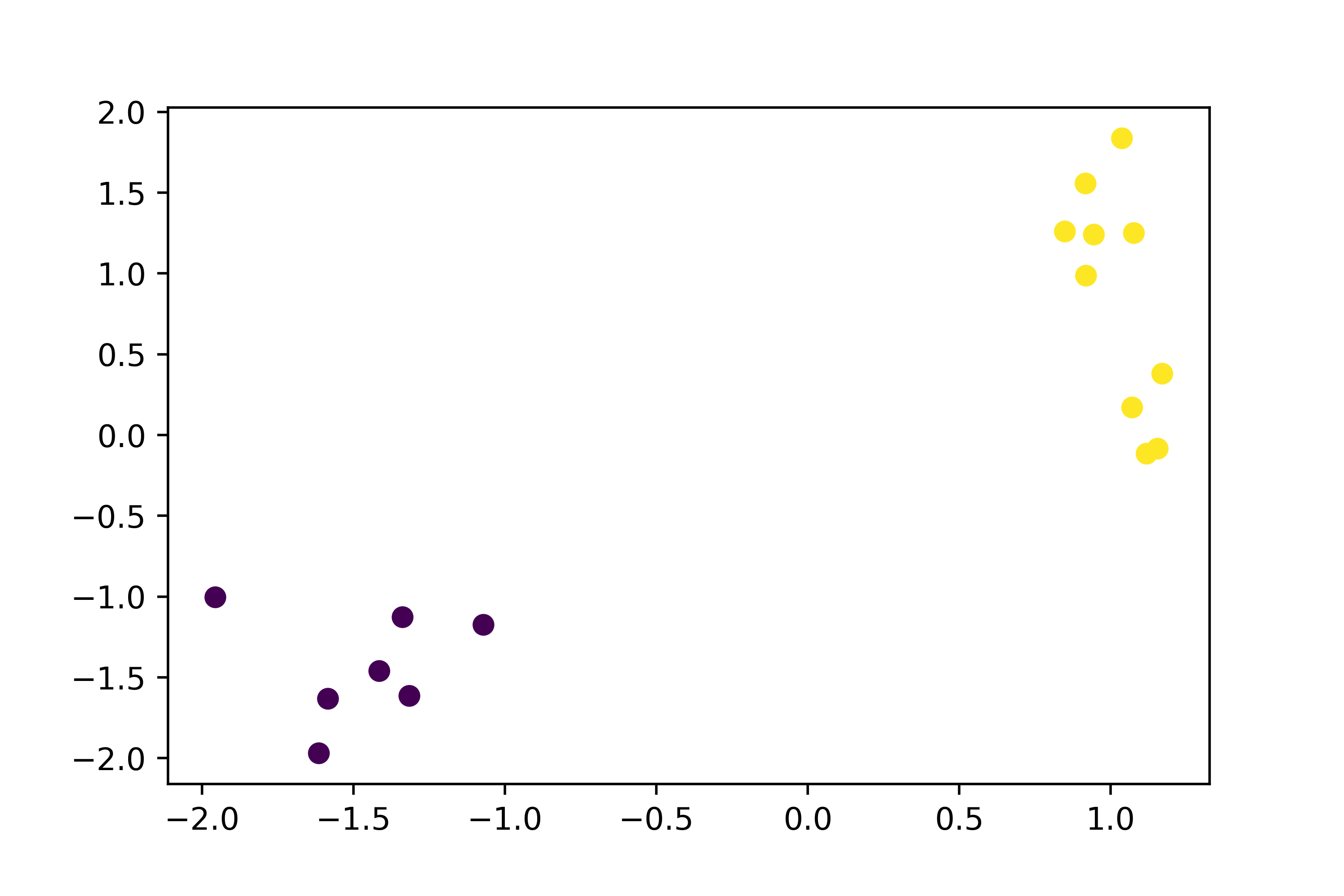}
  \caption{Case 1: random dataset}
  \label{fig:syn_rand_data_1}
\end{subfigure}
\endminipage\hfill
\minipage{0.3\textwidth}
\begin{subfigure}{\textwidth}
  \centering
\includegraphics[width=\linewidth]{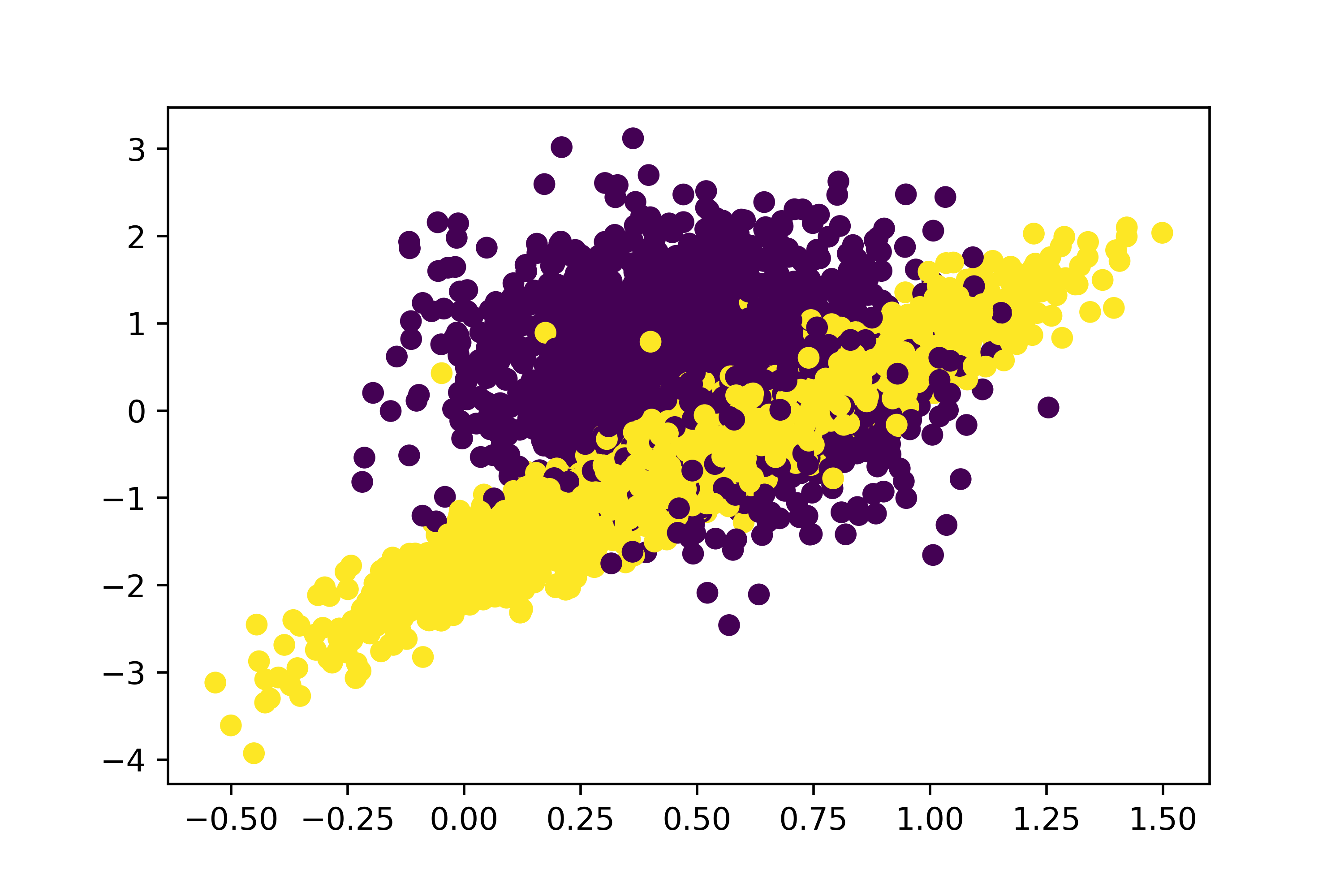}
  \caption{Case 2: original data}
  \label{fig:syn_or_data_2}
\end{subfigure}
\endminipage\hfill
\minipage{0.3\textwidth}
\begin{subfigure}{\textwidth}
  \centering
\includegraphics[width=\linewidth]{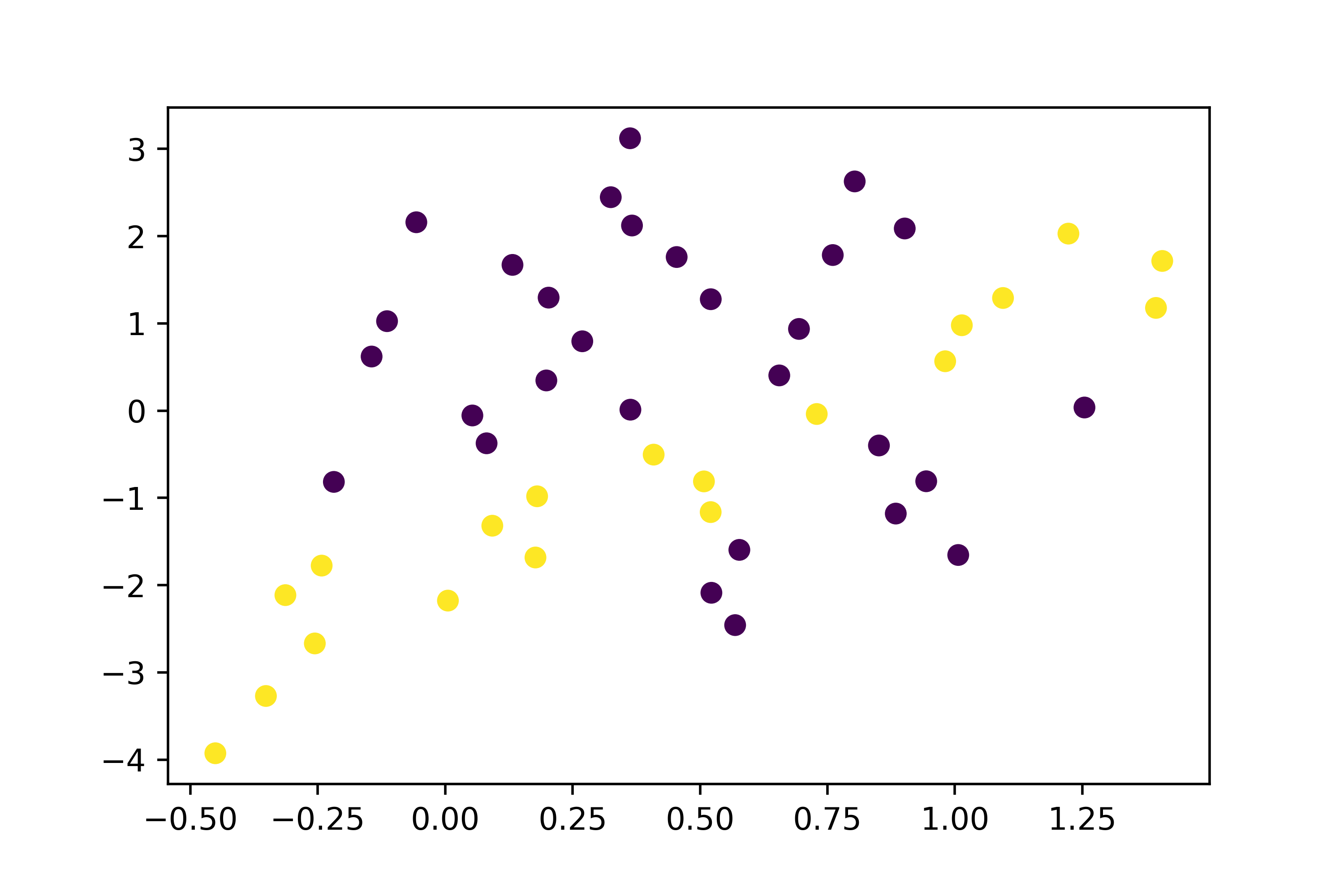}
  \caption{Case 2: dominating 
  dataset}
  \label{fig:syn_dom_data_2}
\end{subfigure}
\endminipage\hfill
\minipage{0.3\textwidth}
\begin{subfigure}{\textwidth}
  \centering
\includegraphics[width=\linewidth]{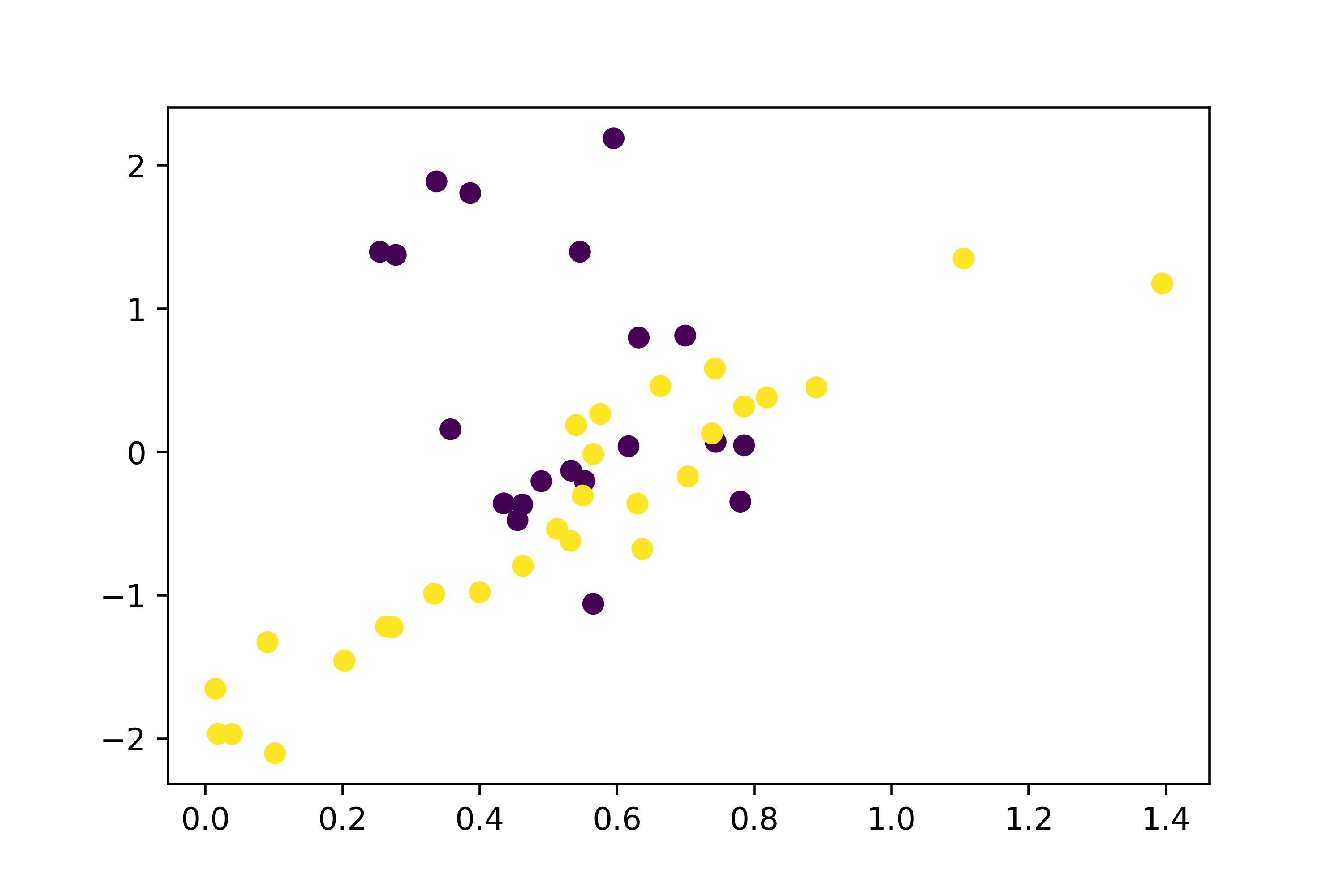}
  \caption{Case 2: random dataset}
  \label{fig:syn_rand_data_2}
\end{subfigure}
\endminipage\hfill
\minipage{0.3\textwidth}
\begin{subfigure}{\textwidth}
  \centering
\includegraphics[width=\linewidth]{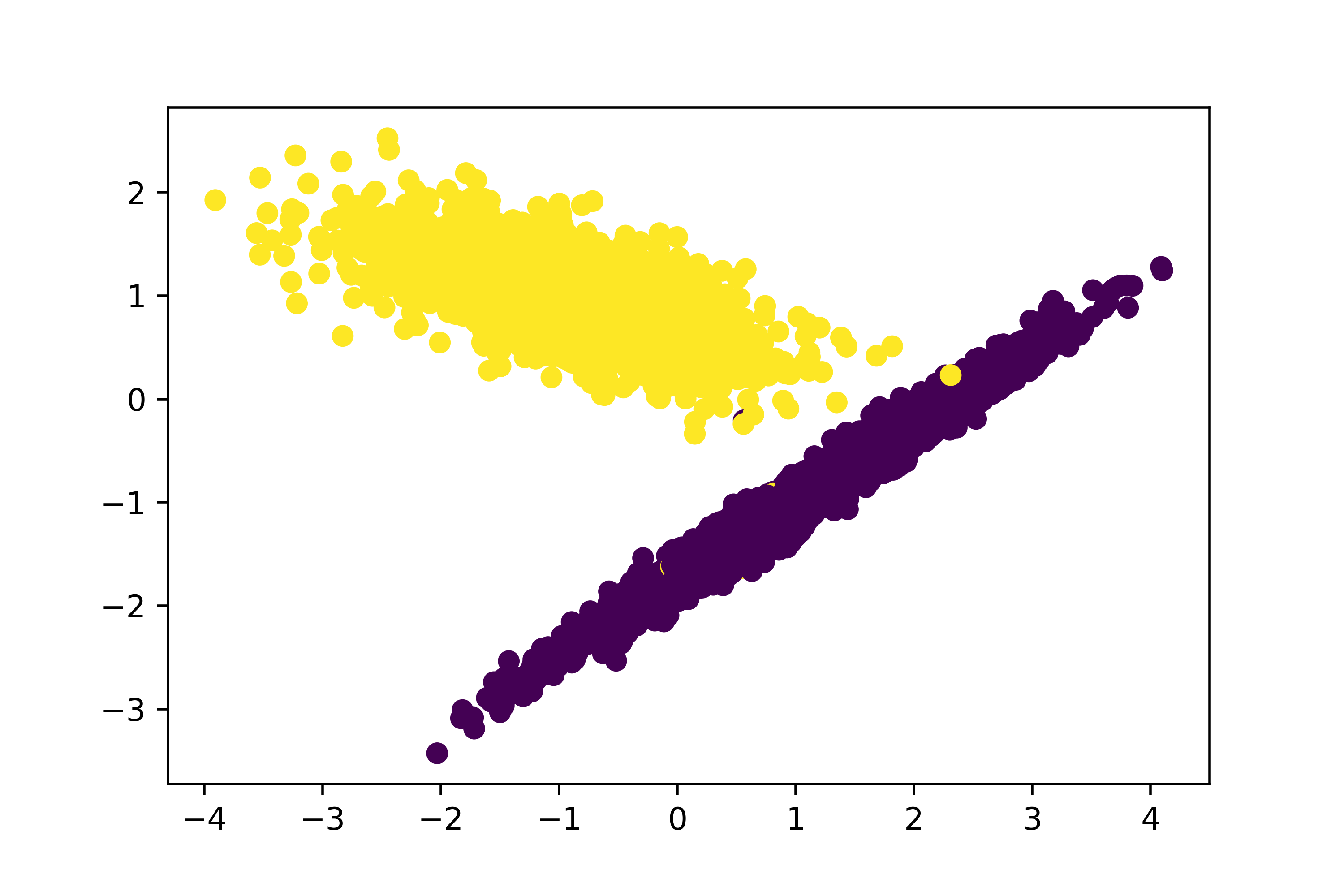}
  \caption{Case 3: original data}
  \label{fig:syn_or_data_3}
\end{subfigure}
\endminipage\hfill
\minipage{0.3\textwidth}
\begin{subfigure}{\textwidth}
  \centering
\includegraphics[width=\linewidth]{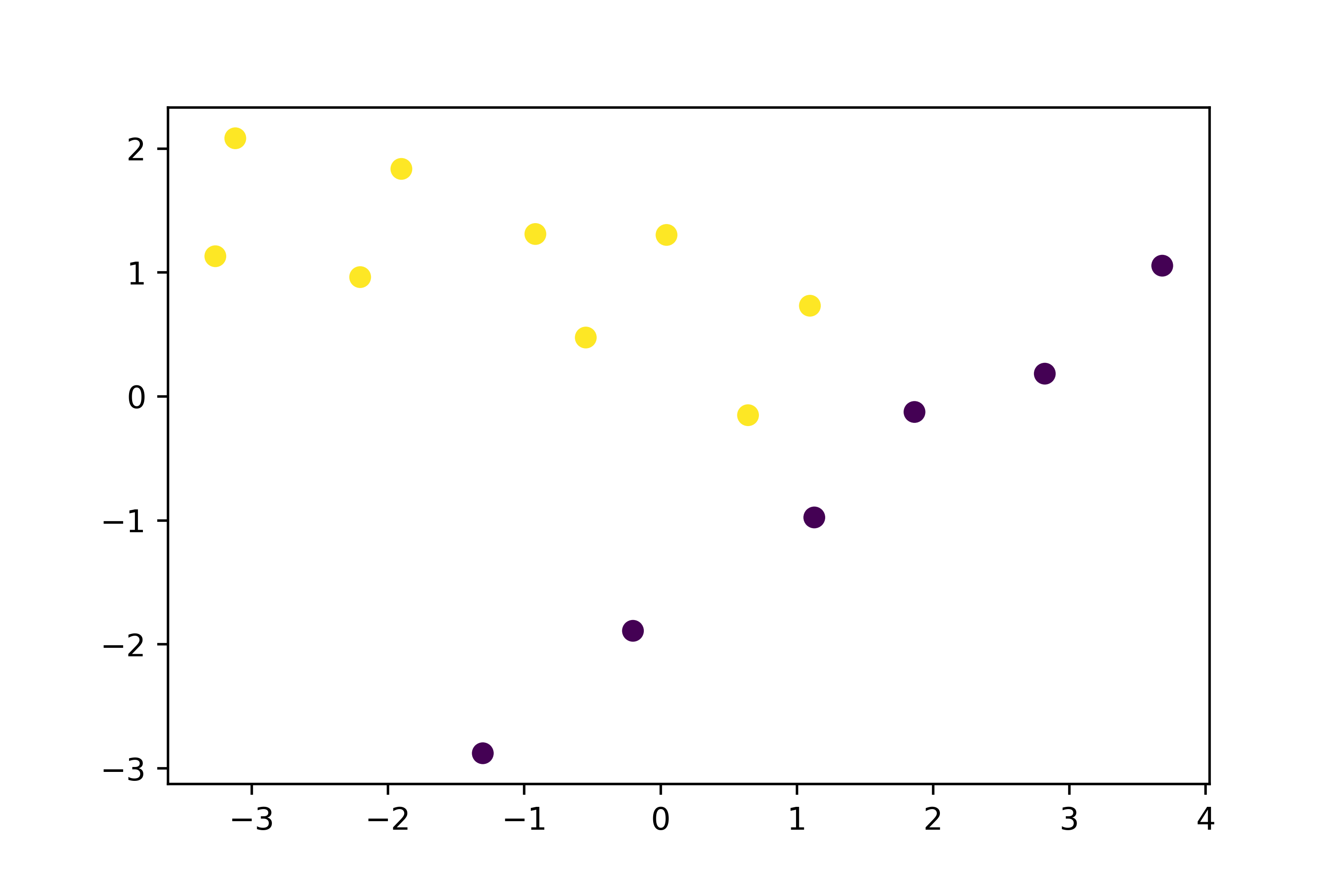}
  \caption{Case 3: dominating 
  dataset}
  \label{fig:syn_dom_data_3}
\end{subfigure}
\endminipage\hfill
\minipage{0.3\textwidth}
\begin{subfigure}{\textwidth}
  \centering
\includegraphics[width=\linewidth]{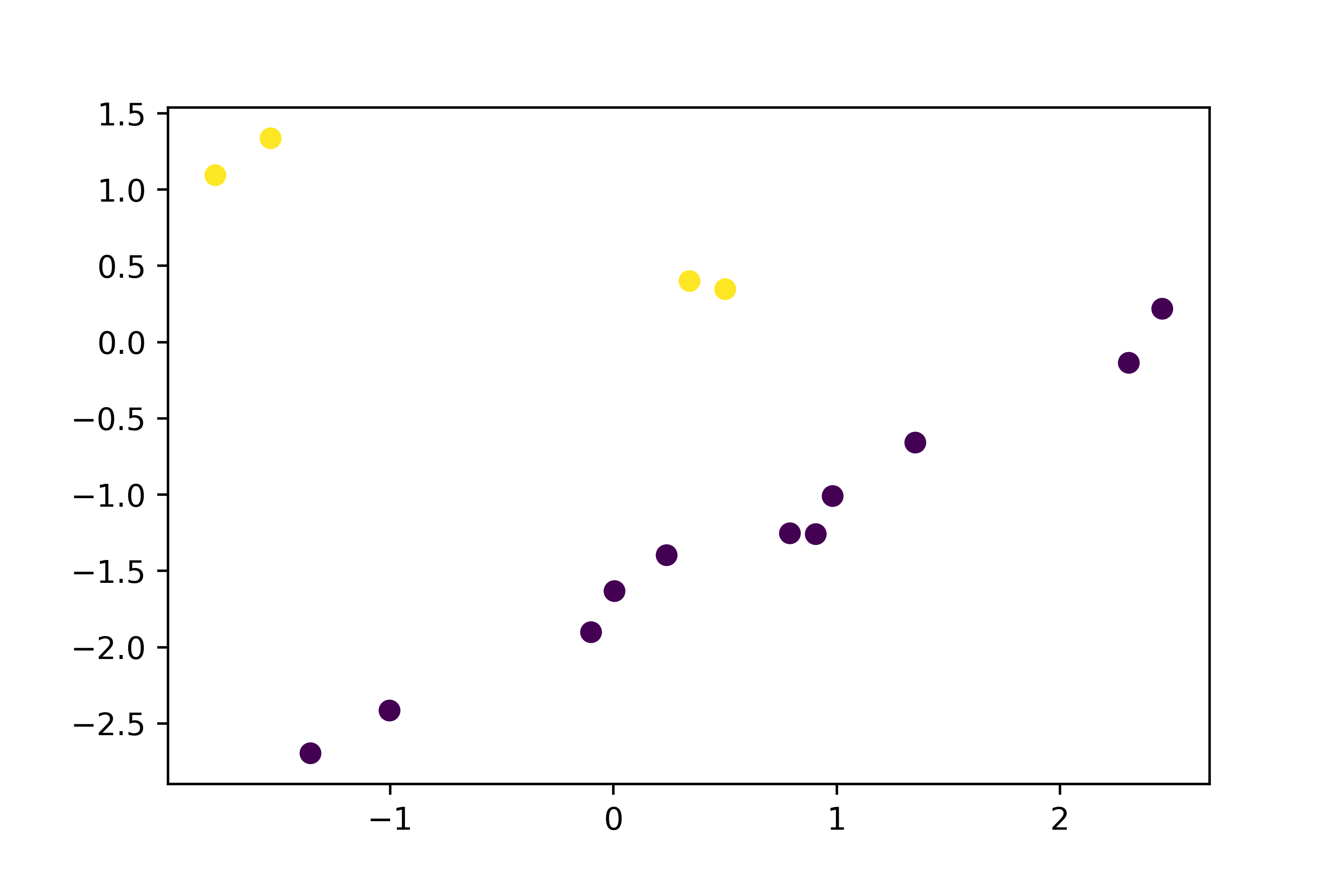}
  \caption{Case 3: random dataset}
  \label{fig:syn_rand_data_3}
\end{subfigure}
\endminipage\hfill
\caption{Different synthetic datasets generated using the Scikit-learn python package implementation. The first column corresponds to original datasets, the second column corresponds to  dominating datasets of the original datasets, and the third column corresponds to  random subsets of the original datasets of the same size as the corresponding dominating set.}
\label{fig:synthetic_perceptron}
\end{figure}

\subsubsection{Synthetic datasets}\label{sec:syn_perceptron}

In this experiment, different datasets were generated using a Scikit-learn python package implementation\footnote{It can be found in \url{https://scikit-learn.org/stable/modules/generated/sklearn.datasets.make_classification.html}}. Roughly speaking, it creates clusters of normally distributed points  in an hypercube and adds some noise.
Specifically, we considered three different situations: (1) distributions without overlapping; (2) distributions with overlapping; and, (3) a dataset with a ``thin" class and a high $\varepsilon$. In the last experiment, we wanted to show that the choice of $\varepsilon$ is important, and that there are cases where representative datasets are not so useful. In all the three cases, the perceptron was trained using the stochastic gradient descent algorithm and the mean squared error as the loss function.

In the first case (see Fig. \ref{fig:syn_or_data_1}), $5000$ points were taken with two clusters well differentiated, i.e. without overlapping between classes. 
The $20\%$ of the points were selected to belong to the test set and the rest of the points constituted the training dataset. 
Then, an $\varepsilon$-representative dataset of the training dataset was computed using Algorithm~\ref{algo} with $\varepsilon=0.8$, obtaining a dominating dataset with just $17$ points. Similarly, $17$ random points were chosen from the training dataset (see Fig. \ref{fig:syn_rand_data_1}). 
Later, a perceptron was trained on each dataset for $20$ epochs and evaluated on the test set. The mean accuracy results after $5$ repetitions were: $0.96$ for the dominating dataset, $0.82$ for the random dataset, and $0.98$ for the training dataset. 
Besides, the random dataset reached very low accuracy in general.
In the second case (see Fig. \ref{fig:syn_or_data_2}), a dataset composed of $5000$ points with overlapping classes was generated. As in the first case, the dataset was split into a training and a test set. 
Then, the $\varepsilon$-representative dataset of the training dataset was computed using Algorithm~\ref{algo}
with $\varepsilon=0.5$, resulting in a dominating dataset of size $22$. After training a perceptron for $20$ epochs and repeating the experiments $5$ times, the mean accuracy values were: $0.73$ for the dominating set; $0.67$ for the random set; and, $0.86$ for the training set. Finally, in the third case (See Fig. \ref{fig:syn_or_data_3}), one of the classes was very ``thin", in the sense that the points were very close to each other displaying a thin line. Therefore, if a ``big" $\varepsilon$ were chosen, that class would be represented by a pointed line as showed in Fig. \ref{fig:syn_dom_data_3} where $\varepsilon=0.8$, reducing the dominating dataset to $15$ points. 
With this example, we wanted to show a case where representative datasets were not so useful. The perceptron was trained for $20$ epochs and the mean accuracy of $5$ repetitions were: $0.72$ for the dominating set; $0.76$ for the random set; and, $0.99$ for the training set. In terms of time, the training for $20$ epochs with the training set took around $20$ seconds and the training with the dominating dataset took half a second. 
The computation of the dominating dataset took around $7$ seconds. 
In Table~\ref{table:synthetic_perceptron}, some evaluation metrics are provided.

\begin{table}[]
\renewcommand{\arraystretch}{2}
\centering
\begin{tabular}{P{1cm}P{2cm}P{1cm}P{1cm}P{1cm}P{1cm}P{1cm}}
\cline{3-7}
                                              &                & Accuracy & Recall & Precission & AUC  & MSE  \\ \hline
\multicolumn{1}{c}{\multirow{3}{*}{Case 1}} & Training dataset   & 0.98     & 0.99   & 0.98       & 0.99 & 0.01 \\ \cline{2-7} 
\multicolumn{1}{c}{}                        & Dominating 
dataset & 0.96     & 0.97   & 0.97       & 0.99 & 0.09 \\ \cline{2-7} 
\multicolumn{1}{c}{}                        & Random dataset     & 0.82     & 0.96   & 0.82       & 0.84 & 0.15 \\ \hline
\multicolumn{1}{c}{\multirow{3}{*}{Case 2}} & Training dataset   & 0.86     & 0.85   & 0.93       & 0.92 & 0.1  \\ \cline{2-7} 
\multicolumn{1}{c}{}                        & Dominating 
dataset & 0.73     & 0.89   & 0.72       & 0.79 & 0.18 \\ \cline{2-7} 
\multicolumn{1}{c}{}                        & Random dataset     & 0.67     & 0.93   & 0.66       & 0.78 & 0.2  \\ \hline
\multicolumn{1}{c}{\multirow{3}{*}{Case 3}}  & Training dataset   & 0.99     & 0.98   & 0.99       & 0.99 & 0.01 \\ \cline{2-7} 
\multicolumn{1}{c}{}                        & Dominating 
dataset & 0.72     & 0.87   & 0.64       & 0.95 & 0.18 \\ \cline{2-7} 
\multicolumn{1}{c}{}                        & Random dataset     & 0.76     & 0.71   & 0.61       & 0.71 & 0.22 \\ \hline
\end{tabular}
\caption{Evaluation metrics on the test set for  a perceptron trained with the  training datasets, the dominating
datasets and the random datasets computed from the synthetic datasets showed in Figure~\ref{fig:synthetic_perceptron}.}
\label{table:synthetic_perceptron}
\end{table}

% Perceptrons, initiated with same weights, were trained with stochastic gradient descent using the different datasets. They were trained along $100$ iterations. Table \ref{table:time} illustrates  that the training time improves when using smaller datasets.
% Fig. \ref{fig:accuracygraph} shows the accuracy of 
% the trained perceptrons measured on the datasets that were used for its training process, illustrating that the training process were successful for all of them. 
% In Fig. \ref{fig:oraccuracygraph}, we can see the accuracy of the trained perceptrons measured on the original dataset. All of them converge to similar accuracy. 
% However, we can see that the dominating dataset has a faster convergence than the other two datasets.

% In Table \ref{table:difErrors}, the means of the error differences obtained after the different training processes are shown. The maximum and minimum error value were $0.38$ and $0.18$, respectively, for the random dataset, and $0.19$ and $0.18$  for the original and the dominating datasets. Therefore, we can conclude that  the dominating dataset generalizes better the original dataset than the random one.

\subsubsection{The Iris Dataset}\label{subsec:exp2}

\begin{figure*}%[!htb]
\minipage{0.32\textwidth}
  \includegraphics[width=\linewidth]{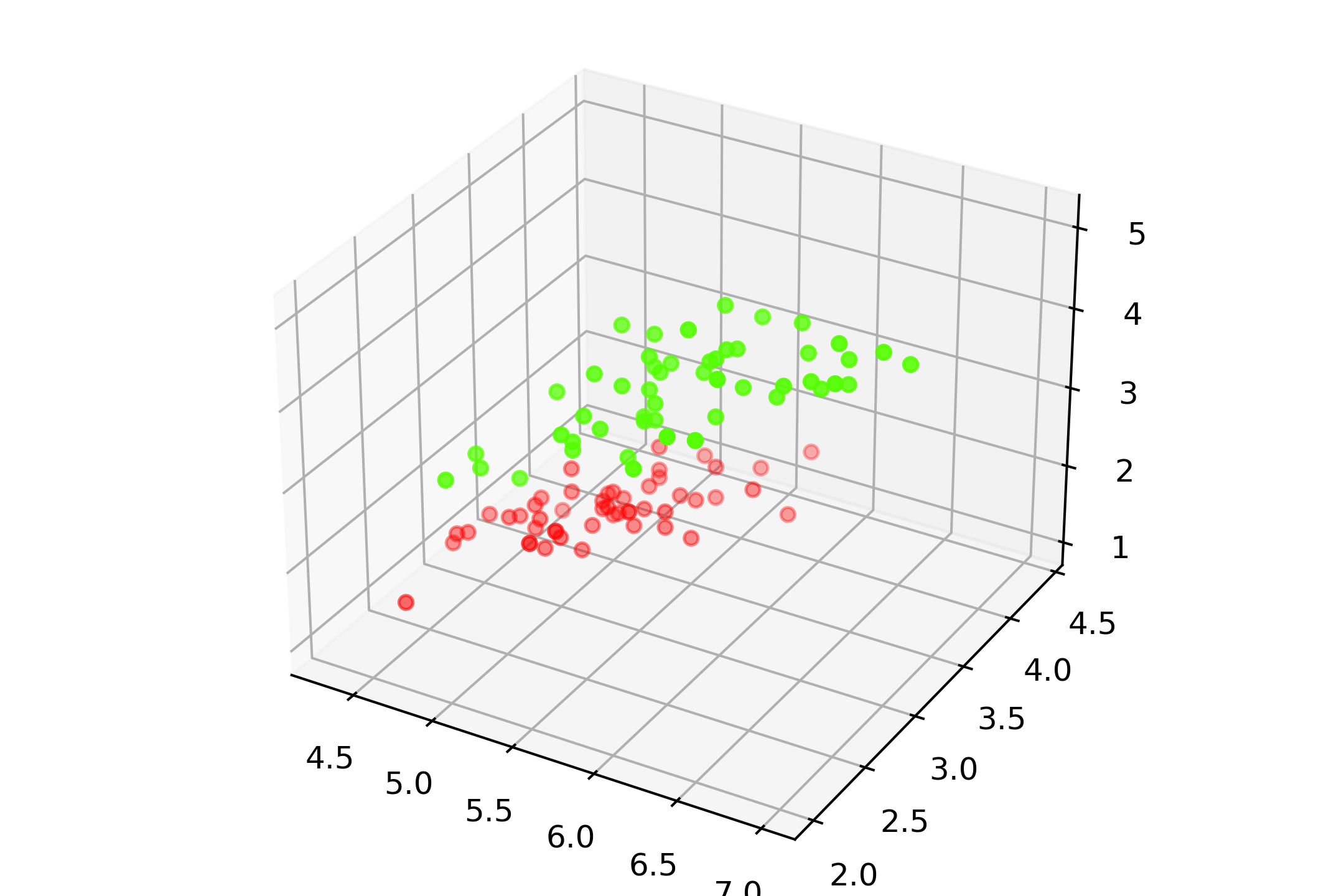}
  \subcaption{Iris dataset}\label{iris_or}
\endminipage\hfill
\minipage{0.32\textwidth}
  \includegraphics[width=\linewidth]{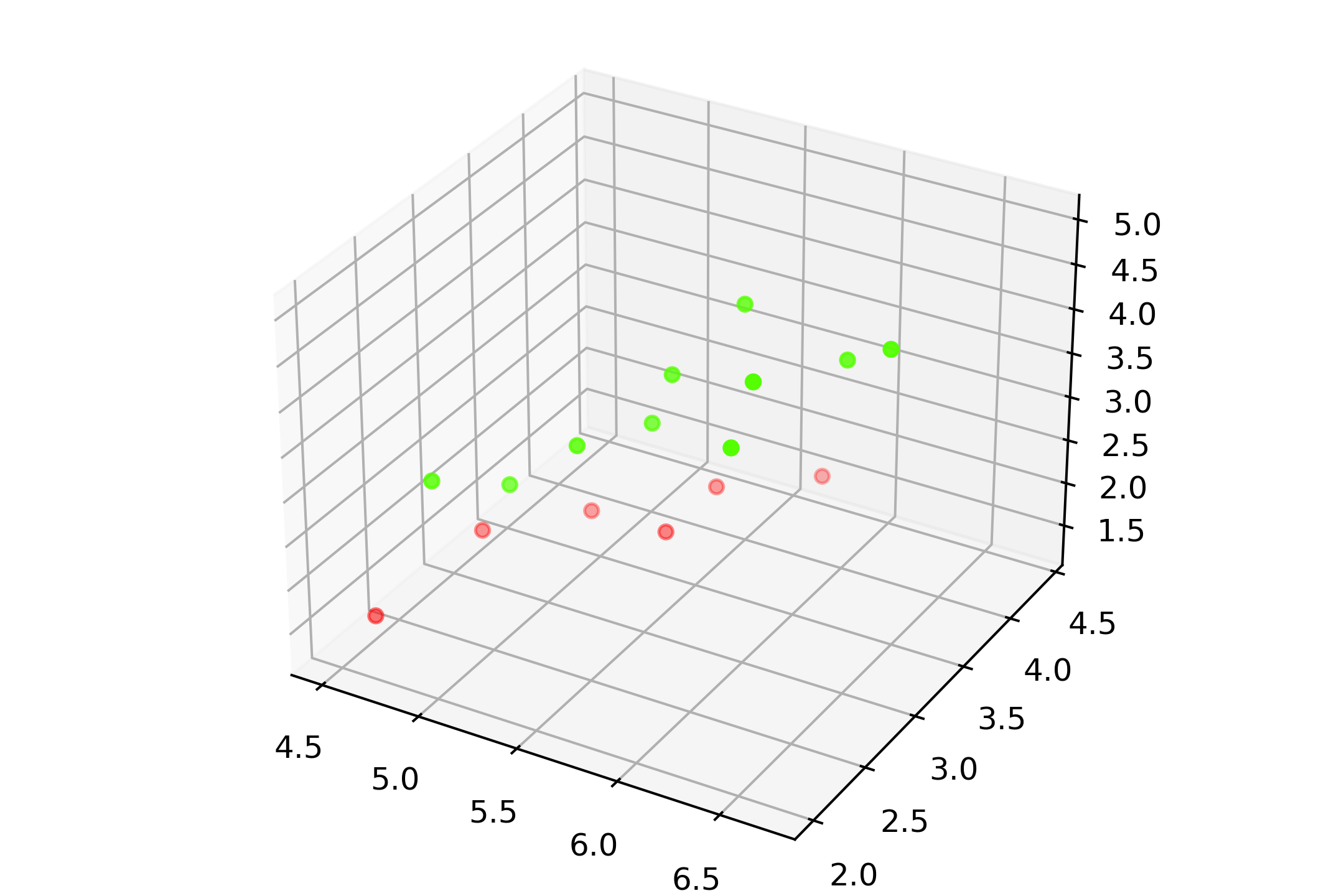}
  \subcaption{Iris dominating dataset}\label{iris_dom}
\endminipage\hfill
\minipage{0.32\textwidth}
  \includegraphics[width=\linewidth]{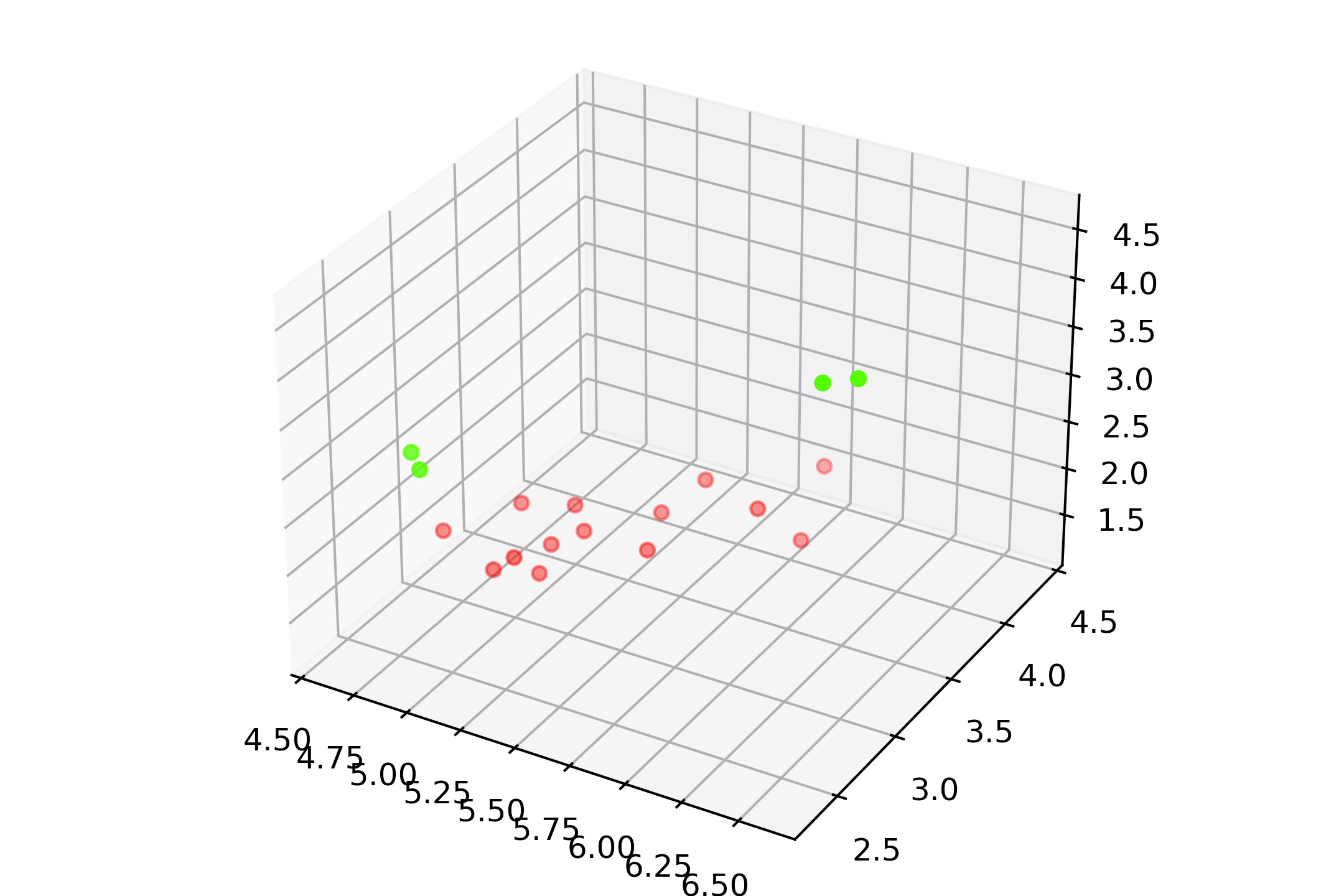}
  \subcaption{Iris random dataset}\label{iris_rand}
\endminipage\hfill
%\minipage{0.32\textwidth}
  %\includegraphics[width=\linewidth]{Figures/tsne_digits.png}
%   \subcaption{T-SNE embedding digits dataset}\label{fig:tsne_or}
% \endminipage\hfill
% \minipage{0.32\textwidth}
%   \includegraphics[width=\linewidth]{Figures/tsne_dom.png}
%   \subcaption{T-SNE embedding digits dominating dataset}\label{fig:tsne_dom}
% \endminipage\hfill
% \minipage{0.32\textwidth}
%   \includegraphics[width=\linewidth]{Figures/tsne_random.png}
%   \subcaption{T-SNE embedding  digits random dataset}\label{fig:tsne_rand}
%\endminipage

\caption{Visualization of the Iris dataset: the original dataset composed of $100$ points, the dominating dataset, and the random dataset composed of $16$ points.}
\end{figure*}

\begin{figure*}[!htb]
\minipage{0.32\textwidth}
  \includegraphics[width=\linewidth]{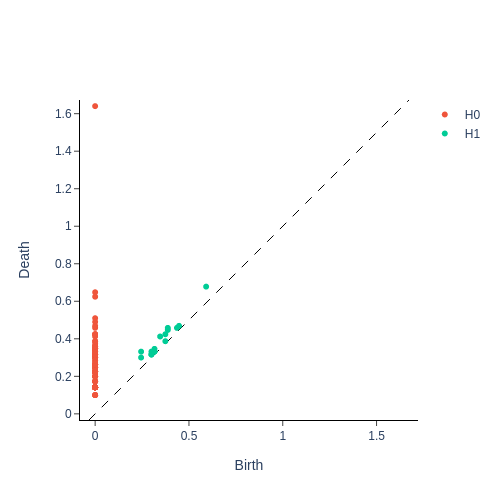}
  \subcaption{{\bf(Iris dataset)} Persistence diagram of the dataset given in Fig. \ref{iris_or}.}\label{fig:pers_iris_or}
\endminipage\hfill
\minipage{0.32\textwidth}
  \includegraphics[width=\linewidth]{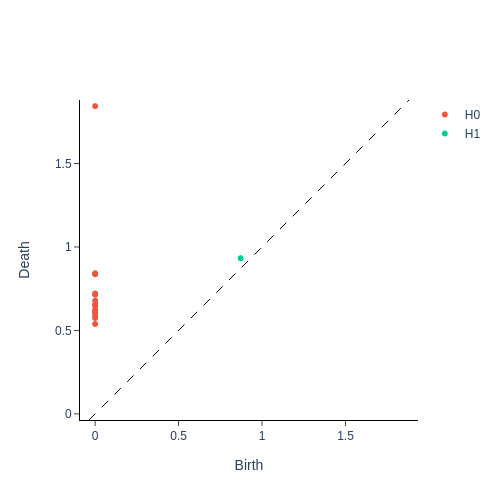}
  \subcaption{{\bf(Iris dataset)} Persistence diagram of the dataset given in Fig. \ref{iris_dom}.}\label{fig:pers_iris_dom}
\endminipage\hfill
\minipage{0.32\textwidth}
  \includegraphics[width=\linewidth]{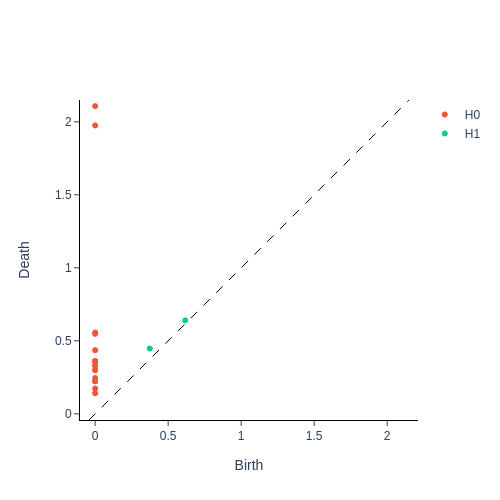}
  \subcaption{{\bf(Iris dataset)} Persistence diagram of the random dataset given in Fig. \ref{iris_rand}.}\label{pers_iris_rand}
\endminipage\hfill
\minipage{0.32\textwidth}
  \includegraphics[width=\linewidth]{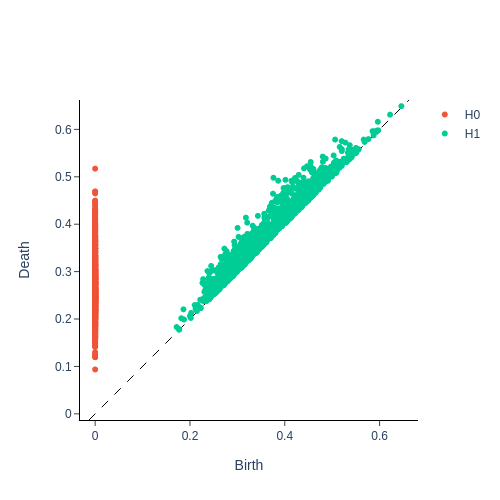}
  \subcaption{{\bf(Digits dataset)} Persistence diagram of the dataset used in Section~\ref{sec:digits}.}\label{fig:pers_digits_or}
\endminipage\hfill
\minipage{0.32\textwidth}
  \includegraphics[width=\linewidth]{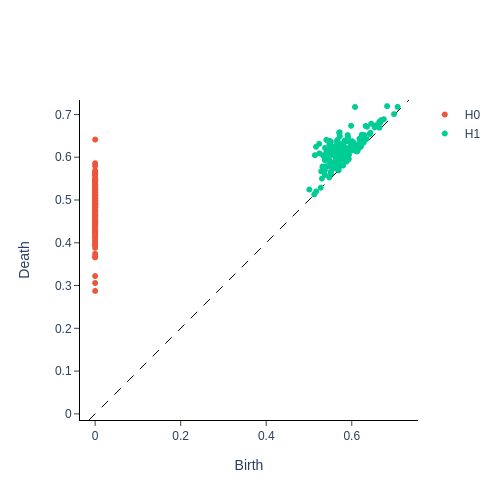}
  \subcaption{{\bf(Digits dataset)} Persistence diagram of the dominating dataset used in Section~\ref{sec:digits}.}\label{fig:pers_digits_dom}
\endminipage\hfill
\minipage{0.32\textwidth}
  \includegraphics[width=\linewidth]{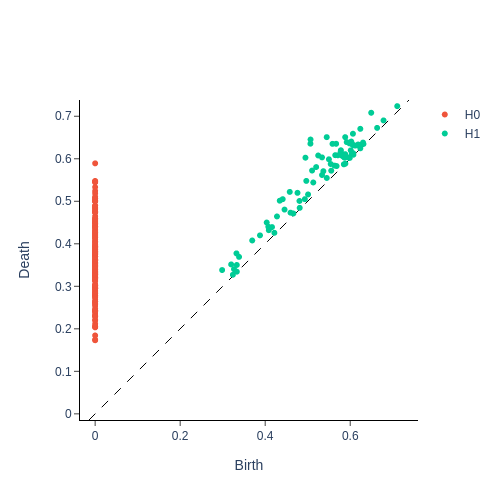}
  \subcaption{{\bf(Digits dataset)} Persistence diagram of the random dataset used in Section~\ref{sec:digits}.}\label{fig:pers_digits_rand}
\endminipage
\caption{Persistence diagrams of the set of points in the datasets used/computed in the experiments.}
\end{figure*}

In this experiment we used the Iris Dataset\footnote{\url{https://archive.ics.uci.edu/ml/datasets/iris}} which corresponds to a classification problem with three classes. It is composed by $150$ 4-dimensional instances. We limited our experiment to two of the three classes, keeping a balanced dataset of $100$ points. Algorithm \ref{algo} was applied to obtain an $\varepsilon$-representative dataset of \edu{$16$} points with $\varepsilon \le 0.5$ (called the dominating dataset). A random dataset extracted from the original dataset with the same number of points than the dominating dataset was also computed for testing. 
These datasets are represented in $\mathbb{R}^3$ in Fig. \ref{iris_or}, Fig. \ref{iris_dom} and Fig. \ref{iris_rand}, respectively. Besides, the associated persistence diagrams are shown in Fig. \ref{fig:pers_iris_or}, \ref{fig:pers_iris_dom} and \ref{pers_iris_rand}.
The Hausdorff and the 0-dimensional bottleneck distances between the original dataset and the dominating and random datasets are given in Table \ref{table:metrics}.

We trained the perceptron with different initial weights and observed that the perceptron trained with the dominating and the original datasets  converged to similar errors. \edu{In Table~\ref{table:difErrors}, the difference between the errors using a fix set of weights for the dominating and the random dataset is provided. In Table~\ref{table:metrics_iris_digits} different metrics were evaluated on the original dataset when training with the original dataset, the dominating set, and the random set. The table shows how the dominating dataset provides better metrics than the random dataset.}
% However,  the perceptron trained with  the random  dataset  converged to a very different error.
%  This can be appreciated in Table \ref{table:difErrors}, where the error differences and the bounds of Lemma \ref{th:expectation} for the random and the dominating datasets are shown. 
%  The maximum and minimum error values were $0.71$ and $0.004$, respectively, for the random dataset, $0.45$ and $0.004$  for the dominating dataset, and $0.5$ and $0.003$ for the original dataset. 
%  These values and the values given in Table \ref{table:difErrors} were obtained after  training the perceptron  $100$ times using different weights. Therefore, we can assert that  the dominating dataset generalizes better the original dataset than the random one.

\begin{table}[]
\renewcommand{\arraystretch}{2}
\centering
\begin{tabular}{P{2cm}P{2cm}P{2cm}P{2cm}P{2cm}}
\hline
Dataset &  & Size & $\frac{1}{2}d_B$ &$d_H$ \\
\hline
   & Original   & 100  & \multicolumn{2}{c}{} \\ \cline{2-5} 
           & Dominating & 16   &      0.11            &   0.58                      \\ \cline{2-5} 
\multirow{-3}{*}{Iris}   & Random     & 16   &  0.5                                   &          1.12               \\ \hline
                                 & Original   & 1797 & \multicolumn{2}{c}{}                        \\ \cline{2-5} 
                                 & Dominating & 173  &  0.09   &    0.29                      \\ \cline{2-5} 
\multirow{-3}{*}{Digits} & Random     & 173  &  0.09          &      0.31                  \\ \hline
\end{tabular}
\caption{The Hausdorff 
and the 0-dimensional bottleneck distances of two different datasets (named iris and digits)
and their corresponding dominating and random datasets.} \label{table:metrics}
\end{table}

\begin{table}[!ht]
\renewcommand{\arraystretch}{2}
\centering
\begin{tabular}{P{2cm}P{2cm}P{4cm}}
\hline
\multicolumn{2}{c}{Datasets}       & $ \|E(w,X)-E(w,\tilde{X})\|$\\ \hline
\multirow{2}{*}{Iris}   & Dominating &    0.05                                                 \\ \cline{2-3} 
                        & Random     &   0.27  \\ \hline
\end{tabular}
\caption{Comparison between the exact error differences computed over the random and the dominating dataset for the Iris classification problem. 
The values correspond to the mean of the exact error differences and the bound  obtained  using $100$ different random weights.
%As the datasets were not $\lambda$-balanced, the bound given in Lemma \ref{th:expectation} was not computed. However, it is experimentally shown that with fix weights, the dominating dataset error values are more similar to the original dataset than in the random dataset case.
}
\label{table:difErrors}
\end{table}

\begin{table}
\renewcommand{\arraystretch}{2}
\centering
\begin{tabular}{P{1cm}P{2cm}P{1cm}P{1cm}P{1cm}P{1cm}P{1cm}}
\cline{3-7}
                                              &                & Accuracy & Recall & Precission & AUC  & MSE  \\ \hline
\multicolumn{1}{c}{\multirow{3}{*}{Iris}} & Original dataset   &  1    &  1  &   1     &0.99  & 0.003 \\ \cline{2-7} 
\multicolumn{1}{c}{}                        & Dominating 
dataset &  0.9    & 1   &  0.9      & 0.9 & 0.11 \\ \cline{2-7} 
\multicolumn{1}{c}{}                        & Random dataset     &   0.6   & 0.2   & 0.2       & 0.36 & 0.39 \\ \hline
\multicolumn{1}{c}{\multirow{3}{*}{Digits}} & Original dataset   &   0.95   & 0.94   &  0.96      & 0.99 & 0.01  \\ \cline{2-7} 
\multicolumn{1}{c}{}                        & Dominating 
dataset &  0.77    &  0.76  & 0.78       & 0.95 & 0.04 \\ \cline{2-7} 
\multicolumn{1}{c}{}                        & Random dataset     & 0.63 &    0.62& 0.64       & 0.89 & 0.06  \\ \hline
\end{tabular}
\caption{The different metrics provided for the iris and the digits dataset experiments are provided. Both are the mean values of $5$ repetitions of the experiments and are evaluated on the full original dataset.}
\label{table:metrics_iris_digits}
\end{table}

\begin{table}[]
\renewcommand{\arraystretch}{2}
\centering
\begin{tabular}{P{2cm}P{2cm}P{2cm}P{2cm}}
\hline
\multicolumn{2}{c}{\multirow{2}{*}{Datasets}} & \multicolumn{2}{c}{Time (s)} \\ \cline{3-4} 
\multicolumn{2}{c}{}                          & Alg. \ref{algo} & Training     \\ \hline
\multirow{2}{*}{Iris}         & Original        & &\edu{3.06}         \\ \cline{2-4} 
                              & Dominating      &           \edu{0.04}    &   \edu{0.8}           \\ \hline
\multirow{2}{*}{Digits}       & Original        & & \edu{65.6}         \\ \cline{2-4} 
                              & Dominating      &   \edu{0.57}             &     \edu{11.78}        \\ \hline
\end{tabular}
\caption{Time (in seconds) required to compute the dominating datasets using Algorithm~\ref{algo} and time (in seconds) required for the training process on the different datasets. 
The training method consists of the gradient descent algorithm. In the case of the digits dataset, a multi-layer neural network was trained for $1000$ epochs using the Adam training algorithm.}
\label{table:time}
\end{table}

\section{The Multi-layer Neural Network Case}\label{sec:experiments}

In this section, we will check, experimentally, the usefulness of representative datasets for more complex neural network architectures. 
Two different experiments were made using synthetic datasets and digits images. 
The implementation of the methodology can be found online in \url{https://github.com/Cimagroup/Experiments-Representative-datasets}.

\subsection{Synthetic Datasets}
This experiment consists of two different binary classification problems from synthetic datasets with $5000$ points. The datasets were split into a training set and a test set with proportions of $80\%$ and $20\%$, respectively. 
Then, a dominating dataset and a random dataset of the same size were computed and used for training a $3\times 12 \times 6 \times 1$ multi-layer neural network. 
The neural network used ReLu activation function in the inner layers and sigmoid function in the output layer, and was trained using stochastic gradient descent and mean squared error as loss function for $20$ epochs. 

\begin{figure}
  \centering
\minipage{0.32\textwidth}
\includegraphics[width=\linewidth]{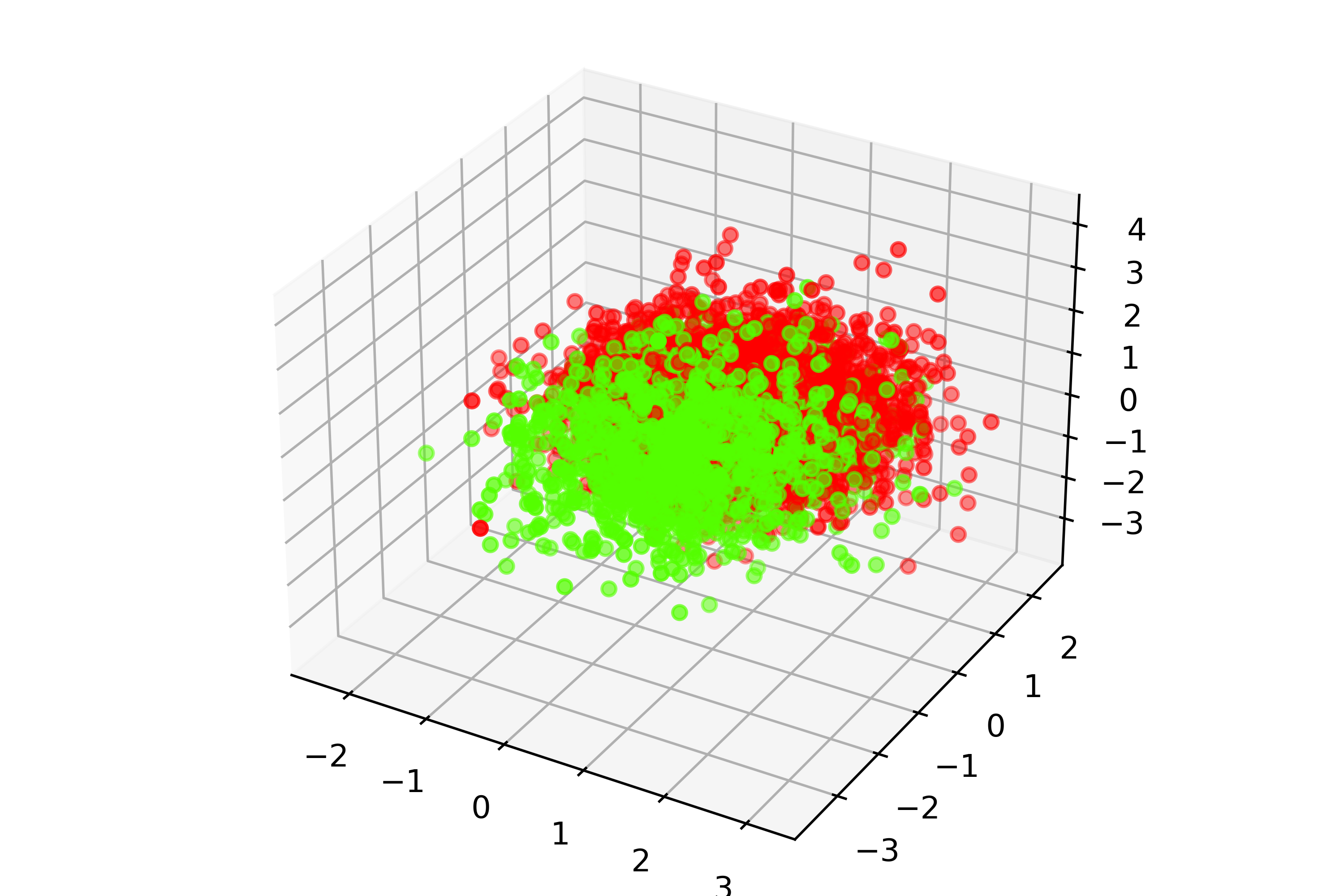}
  \subcaption{Case 1: original dataset}
  \label{fig:nn_syn_or_data_1}
\endminipage\hfill
\minipage{0.32\textwidth}
  \centering
  \includegraphics[width=\linewidth]{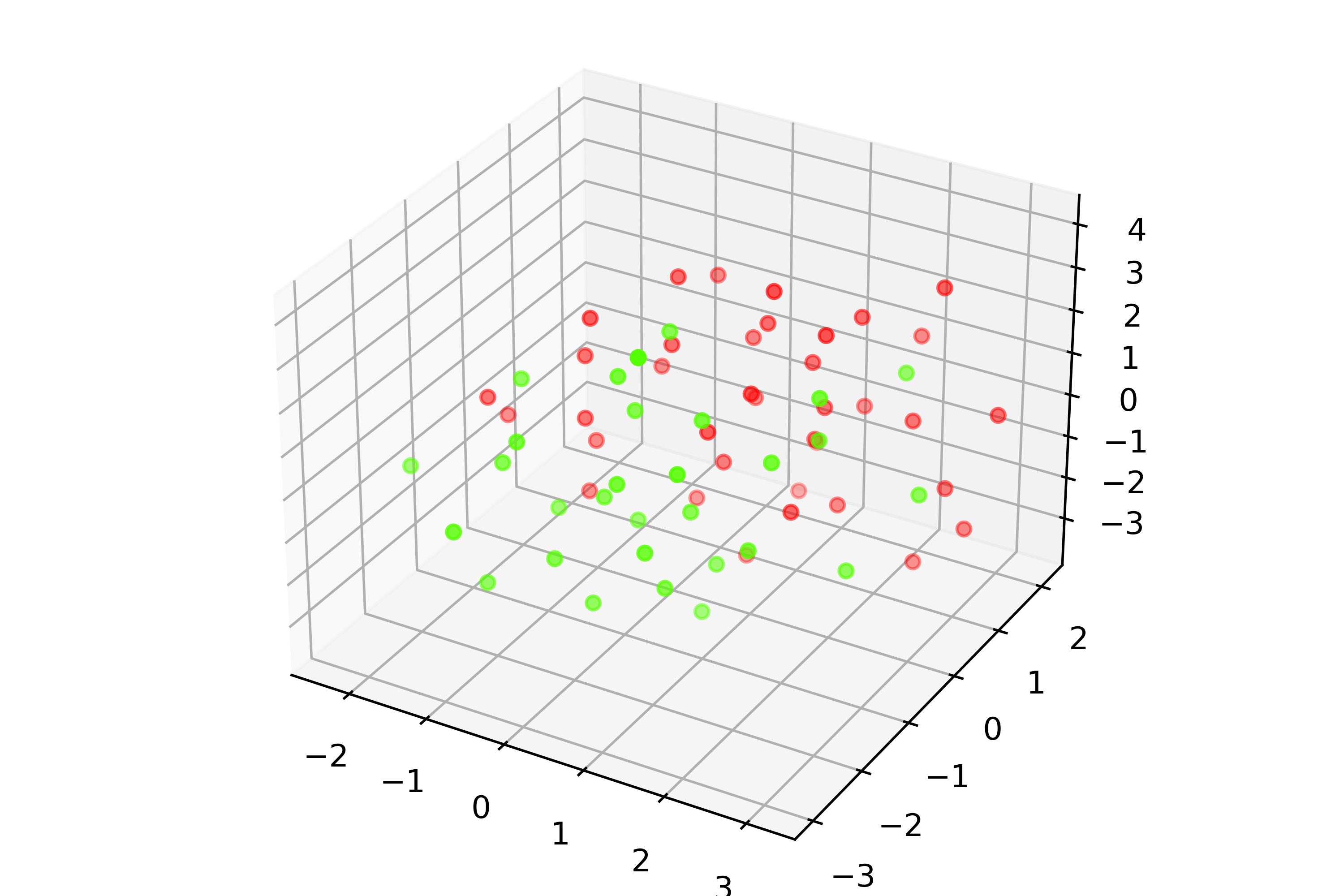}
  \subcaption{Case 1: dominating dataset}
  \label{fig:nn_syn_dom_data_1}
\endminipage\hfill
\minipage{0.32\textwidth}
  \centering
  \includegraphics[width=\linewidth]{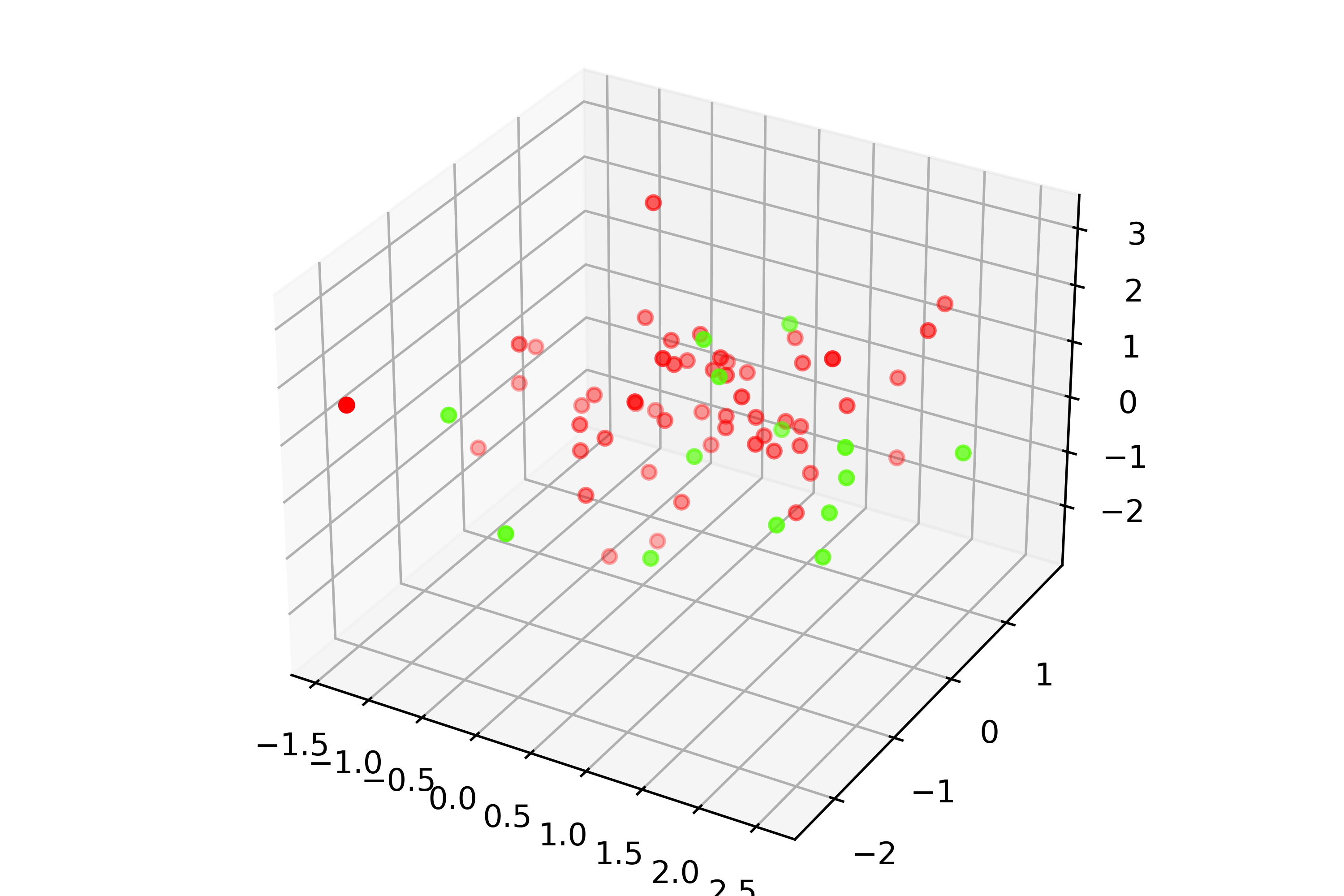}
  \subcaption{Case 1: random dataset}
  \label{fig:nn_syn_rand_data_1}
\endminipage\hfill
\minipage{0.32\textwidth}
  \centering
\includegraphics[width=\linewidth]{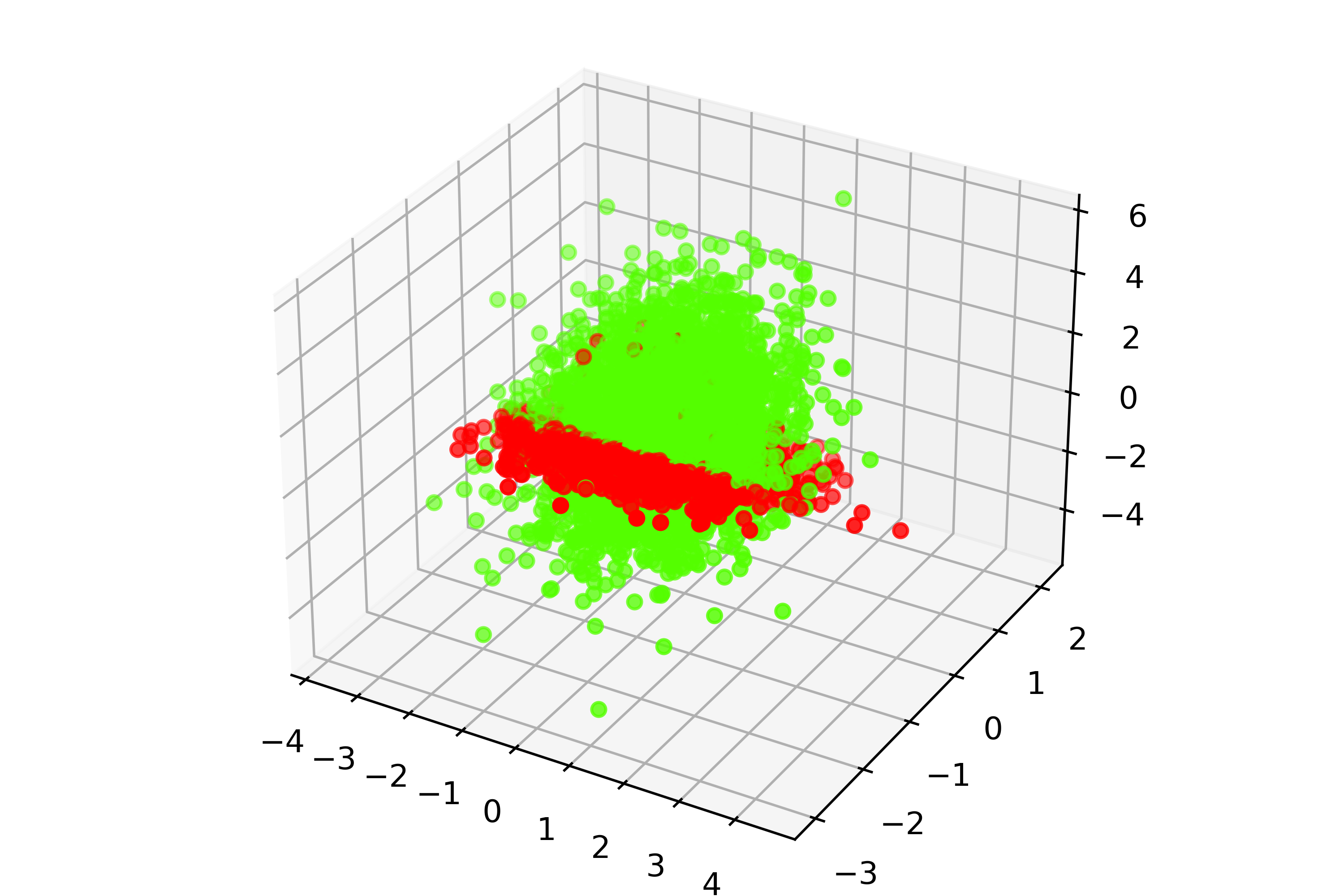}
  \subcaption{Case 2: original dataset}
  \label{fig:nn_syn_or_data_2}
\endminipage\hfill
\minipage{0.32\textwidth}
  \centering
  \includegraphics[width=\linewidth]{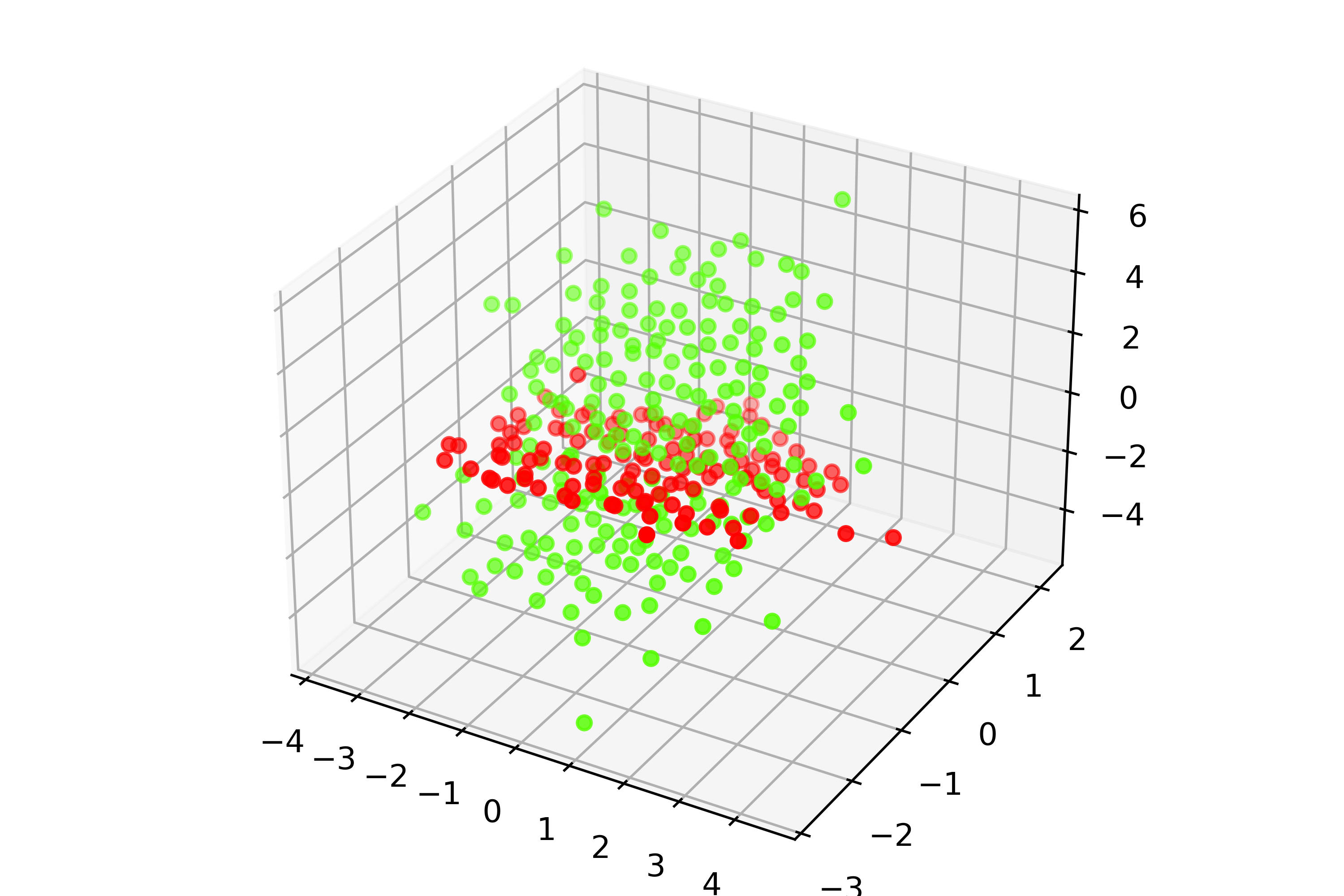}
  \subcaption{Case 2: dominating dataset}
  \label{fig:nn_syn_dom_data_2}
\endminipage\hfill
\minipage{0.32\textwidth}
  \centering
  \includegraphics[width=\linewidth]{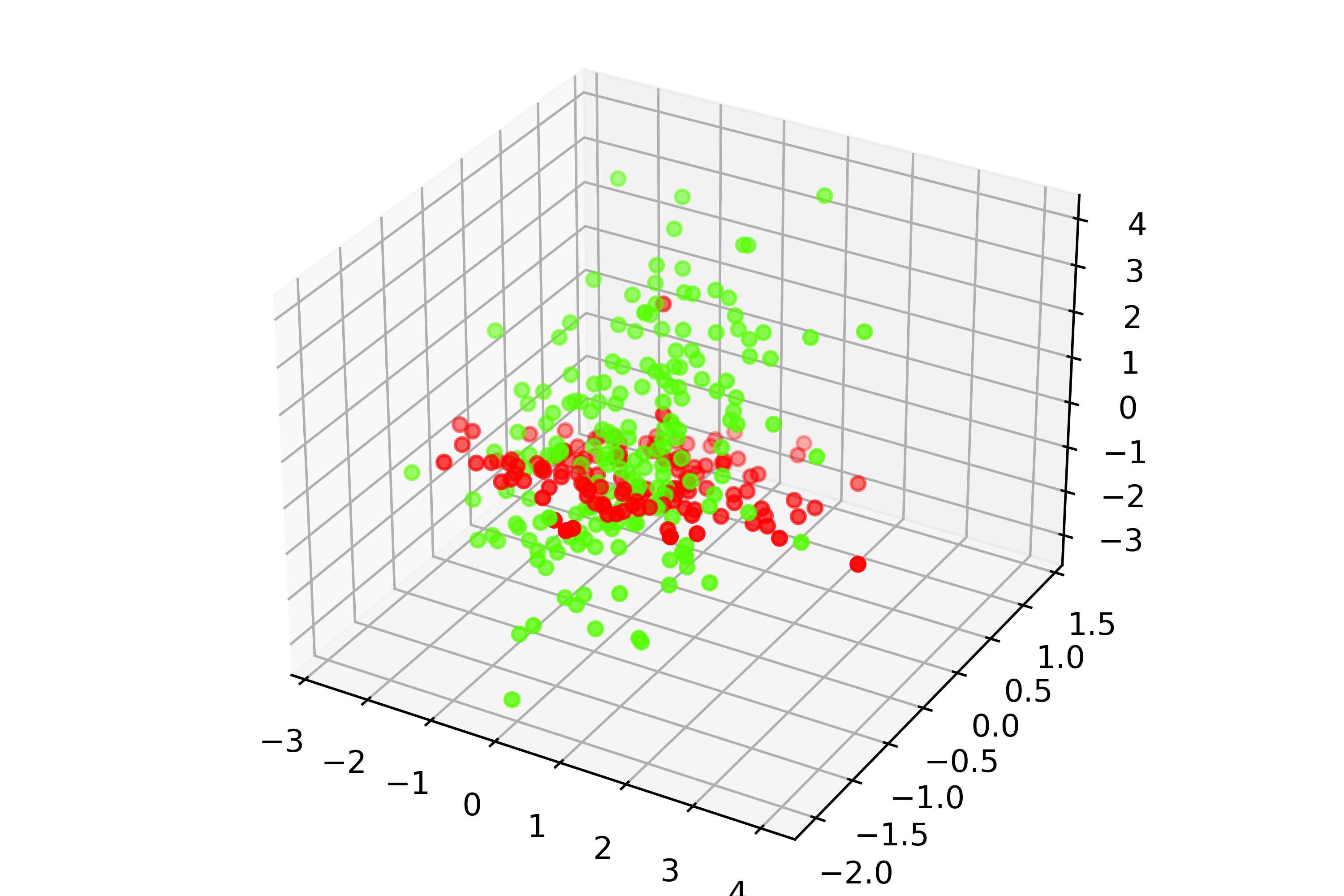}
  \subcaption{Case 2: random dataset}
  \label{fig:nn_syn_rand_data_2}
\endminipage\hfill

\caption{Different synthetic datasets generated using the Scikit-learn python package implementation. 
The first column correspond to the original datasets, the second column are dominating datasets computed from the original dataset, and the third column are random subsets of the original data of the same size as the corresponding dominating set.}
\label{fig:syn_nn}
\end{figure}

In the first case (see Fig.~\ref{fig:nn_syn_or_data_1}), an unbalanced dataset with overlapping was considered, and a $\varepsilon$-dominating dataset was computed with $\varepsilon=0.8$ composed of $67$ points (see Fig.~\ref{fig:nn_syn_dom_data_1}). 
Then, a random dataset with the same size as the dominating dataset was considered. The neural network was trained and the mean accuracy values after $5$ repetitions were: $0.85$ for the dominating set; $0.74$ for the random set; and, $0.86$ for the training set. 
In the second case, a balanced dataset with overlapping was considered (see Fig.~\ref{fig:nn_syn_or_data_2}), and the same process as in the first case was followed but with $\varepsilon=0.3$, obtaining a dominating set of size $319$. 
Then, the mean accuracy values after $5$ repetitions were: $0.92$ for the dominating set; $0.91$ for the random set; and, $0.93$ for the training set. 
In Table~\ref{table:synthetic_nn}, different evaluation metrics on the test set for the two cases are shown.

\begin{table}[]
\renewcommand{\arraystretch}{2}
\centering
\begin{tabular}{P{1cm}P{2cm}P{1cm}P{1cm}P{1cm}P{1cm}P{1cm}}
\cline{3-7}
                                              &                & Accuracy & Recall & Precission & AUC  & MSE  \\ \hline
\multicolumn{1}{c}{\multirow{3}{*}{Case 1}} & Training set   & 0.86     & 0.61   & 0.85       & 0.88 & 0.11 \\ \cline{2-7} 
\multicolumn{1}{c}{}                        & Dominating set & 0.85     & 0.64   & 0.8        & 0.84 & 0.14 \\ \cline{2-7} 
\multicolumn{1}{c}{}                        & Random set     & 0.74     & 0.05   & 0.38       & 0.73 & 0.18 \\ \hline
\multicolumn{1}{c}{\multirow{3}{*}{Case 2}} & Training set   & 0.93     & 0.95   & 0.93       & 0.98 & 0.05 \\ \cline{2-7} 
\multicolumn{1}{c}{}                        & Dominating set & 0.92     & 0.96   & 0.92       & 0.98 & 0.07 \\ \cline{2-7} 
\multicolumn{1}{c}{}                        & Random set     & 0.91     & 0.94   & 0.92       & 0.98 & 0.09 \\ \hline
\end{tabular}
\caption{Evaluation metrics on the test set for the training of a multi-layer neural network with the training set, the dominating set and the random set of the synthetic dataset experiment.}
\label{table:synthetic_nn}
\end{table}

\subsection{The Digits Dataset}\label{sec:digits}

\begin{figure}[!htb]
\minipage{0.2\textwidth}
  \includegraphics[width=\linewidth]{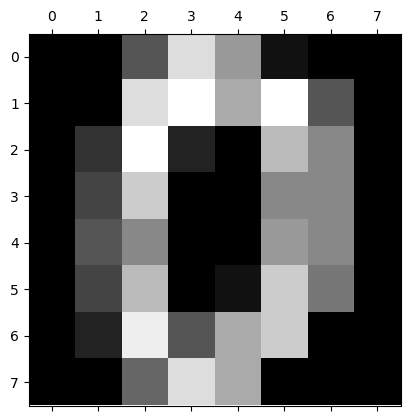}
\endminipage\hfill
\minipage{0.2\textwidth}
  \includegraphics[width=\linewidth]{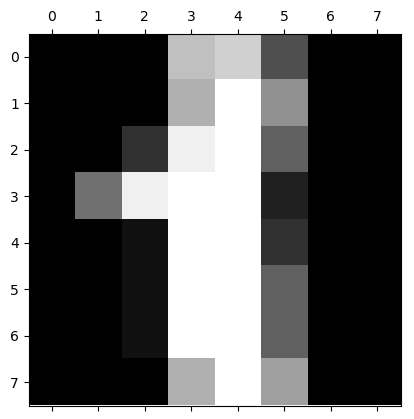}
\endminipage\hfill
\minipage{0.2\textwidth}
  \includegraphics[width=\linewidth]{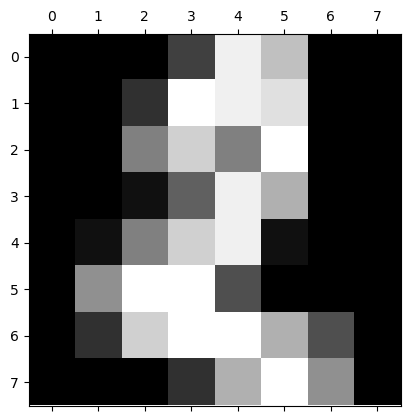}
\endminipage\hfill
\minipage{0.2\textwidth}
  \includegraphics[width=\linewidth]{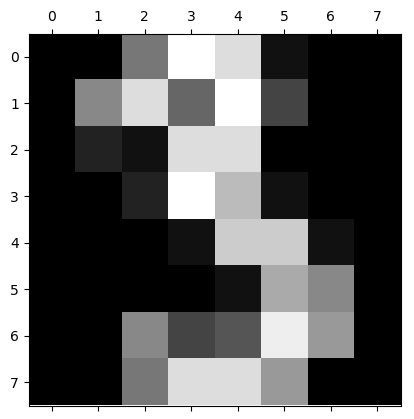}
\endminipage\hfill
\minipage{0.2\textwidth}
  \includegraphics[width=\linewidth]{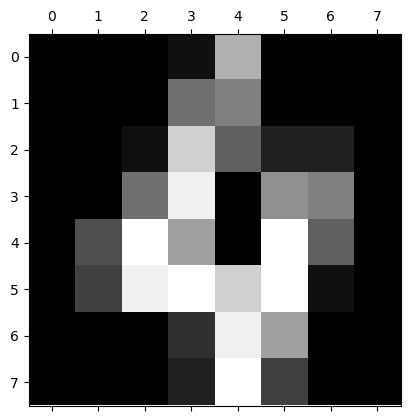}
\endminipage\hfill
\minipage{0.2\textwidth}
  \includegraphics[width=\linewidth]{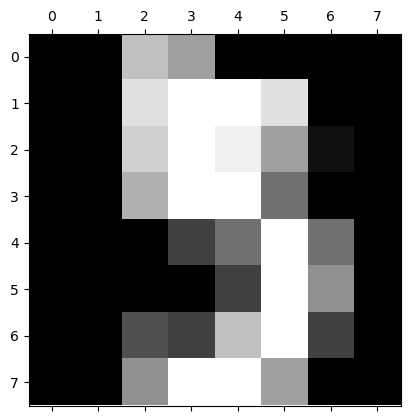}
\endminipage\hfill
\minipage{0.2\textwidth}
  \includegraphics[width=\linewidth]{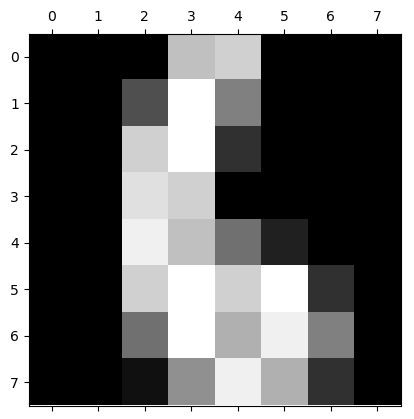}
\endminipage\hfill
\minipage{0.2\textwidth}
  \includegraphics[width=\linewidth]{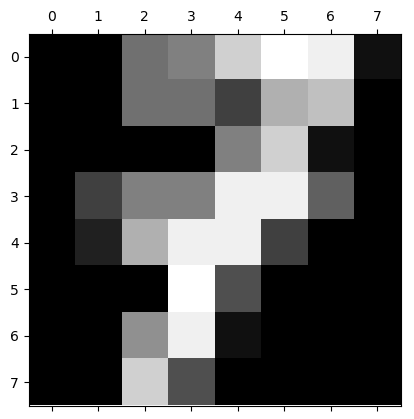}
\endminipage\hfill
\minipage{0.2\textwidth}
  \includegraphics[width=\linewidth]{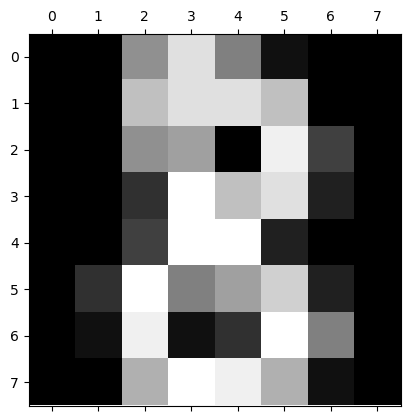}
\endminipage\hfill
\minipage{0.2\textwidth}
  \includegraphics[width=\linewidth]{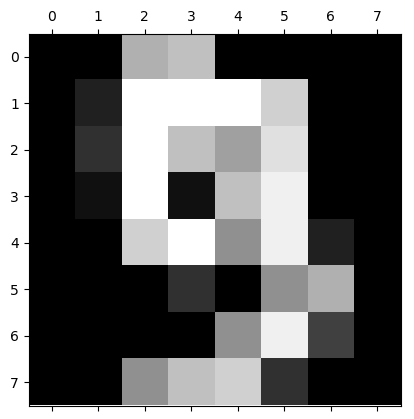}
\endminipage
\caption{{\bf(Digits dataset)} Example of an image of each class of the digits dataset. They are $32\times 32$ arrays in gray scale.}
\label{fig:digits}
\end{figure}

The dataset\footnote{\url{https://scikit-learn.org/stable/auto_examples/datasets/plot_digits_last_image.html}} we will use in this experiment consists of images of digits classified in  $10$ different classes corresponding to digits from $0$ to $9$. 
An example of an image of each class can be seen in Fig. \ref{fig:digits}. The dataset is composed by $1797$ 64-dimensional instances. Algorithm \ref{algo} was applied with $\varepsilon=0.2$ to obtain a dominating dataset of size $173$. 
%In Fig. \ref{fig:tsne_or},  Fig. \ref{fig:tsne_dom} and Fig. \ref{fig:tsne_rand} a 2-dimensional projection of these datasets obtained using T-SNE embedding (see \cite{vanDerMaaten2008}) are shown.
The corresponding persistence diagrams can be seen in Fig. \ref{fig:pers_digits_or}, Fig. \ref{fig:pers_digits_dom} and Fig.  \ref{fig:pers_digits_rand}.  The Hausdorff  and the bottleneck distances are shown in Table \ref{table:metrics}.
In this case, we used a neural network with $64\times 400 \times 300 \times 800 \times 300 \times 10$ neurons with sigmoid activation function in the hidden layers and softmax activation function in the output layer. The neural network was trained using Adam algorithm and categorical cross entropy as loss function for $1000$ epochs. It was launched $5$ times for the dominating dataset and a random dataset of the same size. \edu{The mean accuracy values of different metrics for the $5$ repetitions when training the neural network with the three different datasets and evaluated on the original dataset are shown in Table~\ref{table:metrics_iris_digits} where the dominating dataset outperforms the random dataset. Finally, in Table~\ref{table:diff_eps_digits}, different values for $\varepsilon$ were used to compute the size of the dominating dataset and its accuracy on itself and on the original dataset.}

\begin{table}[ht!]
\renewcommand{\arraystretch}{1.5}
\centering
\begin{tabular}{lrcc}
 $\varepsilon$ & Size & Accuracy evaluated on & Accuracy evaluated on  \\
  &  & the original dataset & the dominating dataset
 \\\hline
      $0.1$       & 1363   &   0.95     & 0.95 \\\hline
     $0.15$       & 531  &   0.91     &  0.98  \\\hline
     $0.18$       & 283    &  0.82    & 0.99    \\
     \hline             
\end{tabular}
\caption{Mean accuracy values after $5$ repetitions of the neural network trained on the dominating dataset and evaluated on both, the dominating dataset and the original dataset for the Digits classification problem.}\label{table:diff_eps_digits}
\end{table}

\section{Conclusions and future work}\label{sec:conclusions}

The success of practical applications and the availability of new hardware (e.g., GPUs \cite{1707.03750} and TPUs \cite{1704.04760}) have led to focus neural network research on the development of new architectures rather than on theoretical issues. 
Such new architectures are one of the pillars of future research in neural networks, but a deeper understanding of the data structure is also necessary for field development, such as new discoveries on adversarial examples \cite{1712.07107} have shown, or the one given in \cite{Forgetting}, where  the redundancy of several datasets is empirically shown.

In this paper, we propose the use of {\it representative datasets} as a new approach to reduce learning time in neural networks based on the topological structure of the dataset. 
Specifically, we  have defined representative datasets using a notion of nearness that has the Gromov-Hausdorff distance as the lower bound. Nevertheless, the bottleneck distance of persistence diagrams (which is a lower bound of the Gromov-Hausdorff distance) is used  to measure the representativeness of the dataset since it is computationally less expensive.  
Furthermore, the agreement between the provided theoretical results and the experiments supports that representative datasets can be a good approach to reach an efficient ``summarization'' of a dataset to train a neural network. First, in the case of gradient descent, the training process is significantly faster when using a representative dataset. 

Planned future work is to provide more experiments using high dimensional real data and different reduction algorithms. Furthermore, we plan to formally prove that the proposed approach can be extended to other neural network architectures and  training algorithms.

\bibliographystyle{plainnat}
\bibliography{biblio}% common bib file

\begin{thebibliography}{27}
\providecommand{\natexlab}[1]{#1}
\providecommand{\url}[1]{\texttt{#1}}
\expandafter\ifx\csname urlstyle\endcsname\relax
  \providecommand{\doi}[1]{doi: #1}\else
  \providecommand{\doi}{doi: \begingroup \urlstyle{rm}\Url}\fi

\bibitem[Biswas and Blanco{-}Medina(2021)]{DBLP:journals/corr/abs-2108-11821}
Rubel Biswas and Pablo Blanco{-}Medina.
\newblock State of the art: Face recognition, 2021.
\newblock URL \url{https://arxiv.org/abs/2108.11821}.

\bibitem[Brown et~al.(2020)Brown, Mann, Ryder, Subbiah, Kaplan, Dhariwal,
  Neelakantan, Shyam, Sastry, Askell, Agarwal, Herbert{-}Voss, Krueger,
  Henighan, Child, Ramesh, Ziegler, Wu, Winter, Hesse, Chen, Sigler, Litwin,
  Gray, Chess, Clark, Berner, McCandlish, Radford, Sutskever, and
  Amodei]{DBLP:journals/corr/abs-2005-14165}
Tom~B. Brown, Benjamin Mann, Nick Ryder, Melanie Subbiah, Jared Kaplan,
  Prafulla Dhariwal, Arvind Neelakantan, Pranav Shyam, Girish Sastry, Amanda
  Askell, Sandhini Agarwal, Ariel Herbert{-}Voss, Gretchen Krueger, Tom
  Henighan, Rewon Child, Aditya Ramesh, Daniel~M. Ziegler, Jeffrey Wu, Clemens
  Winter, Christopher Hesse, Mark Chen, Eric Sigler, Mateusz Litwin, Scott
  Gray, Benjamin Chess, Jack Clark, Christopher Berner, Sam McCandlish, Alec
  Radford, Ilya Sutskever, and Dario Amodei.
\newblock Language models are few-shot learners.
\newblock \emph{CoRR}, abs/2005.14165, 2020.

\bibitem[Chazal et~al.(2009)Chazal, Cohen-Steiner, Guibas, M\'emoli, and
  Oudot]{chazalsignature}
F.~Chazal, D.~Cohen-Steiner, L.~J. Guibas, F.~M\'emoli, and S.~Y. Oudot.
\newblock {Gromov}-{Hausdorff} stable signatures for shapes using persistence.
\newblock \emph{Computer Graphics Forum (proc. SGP 2009)}, pages 1393--1403,
  2009.

\bibitem[Chazal et~al.(2014)Chazal, de~Silva, and Oudot]{Chazal2014}
Fr{\'e}d{\'e}ric Chazal, Vin de~Silva, and Steve Oudot.
\newblock Persistence stability for geometric complexes.
\newblock \emph{Geometriae Dedicata}, 173\penalty0 (1):\penalty0 193--214, Dec
  2014.
\newblock ISSN 1572-9168.
\newblock \doi{10.1007/s10711-013-9937-z}.

\bibitem[Dey and Salem(2017)]{1701.05923}
Rahul Dey and Fathi~M. Salem.
\newblock Gate-variants of gated recurrent unit (gru) neural networks.
\newblock In \emph{2017 IEEE 60th International Midwest Symposium on Circuits
  and Systems (MWSCAS)}, pages 1597--1600, 2017.
\newblock \doi{10.1109/MWSCAS.2017.8053243}.

\bibitem[Edelsbrunner and Harer(2010)]{Edelsbrunner10}
Herbert Edelsbrunner and John~L. Harer.
\newblock \emph{Computational Topology, An Introduction}.
\newblock American Mathematical Society, RI, 2010.

\bibitem[Gonzalez-Diaz et~al.(2018)Gonzalez-Diaz, Paluzo-Hidalgo, and
  Gutierrez-Naranjo]{ENDM2737}
Rocio Gonzalez-Diaz, Eduardo Paluzo-Hidalgo, and Miguel~A. Gutierrez-Naranjo.
\newblock Representative datasets for neural networks.
\newblock \emph{Electronic Notes in Discrete Mathematics}, 68:\penalty0 89--94,
  2018.
\newblock ISSN 1571-0653.
\newblock \doi{10.1016/j.endm.2018.06.016}.
\newblock Discrete Mathematics Days 2018.

\bibitem[Goodfellow et~al.(2016)Goodfellow, Bengio, and
  Courville]{Goodfellow-et-al-2016}
Ian Goodfellow, Yoshua Bengio, and Aaron Courville.
\newblock \emph{Deep Learning}.
\newblock MIT Press, MA, 2016.
\newblock \url{http://www.deeplearningbook.org}.

\bibitem[Gu et~al.(2017)Gu, Liu, Zhou, and Wang]{1707.03750}
Jiazhen Gu, Huan Liu, Yangfan Zhou, and Xin Wang.
\newblock Deepprof: Performance analysis for deep learning applications via
  mining gpu execution patterns, 2017.
\newblock URL \url{https://arxiv.org/abs/1707.03750}.

\bibitem[Hausmann(2016)]{Hausmann1994}
Jean-Claude Hausmann.
\newblock \emph{On the Vietoris-Rips complexes and a Cohomology Theory for
  metric spaces}, pages 175--188.
\newblock Princeton University Press, NJ, 2016.
\newblock \doi{doi:10.1515/9781400882588-013}.

\bibitem[Haykin(2009)]{haykin2009neural}
Simon~S. Haykin.
\newblock \emph{Neural networks and learning machines}.
\newblock Pearson Education, Upper Saddle River, NJ, third edition, 2009.

\bibitem[He et~al.(2016)He, Zhang, Ren, and Sun]{1512.03385}
Kaiming He, Xiangyu Zhang, Shaoqing Ren, and Jian Sun.
\newblock Deep residual learning for image recognition.
\newblock In \emph{2016 IEEE Conference on Computer Vision and Pattern
  Recognition (CVPR)}, pages 770--778, 2016.
\newblock \doi{10.1109/CVPR.2016.90}.

\bibitem[Heck and Salem(2017)]{1701.03452}
Joel~C. Heck and Fathi~M. Salem.
\newblock Simplified minimal gated unit variations for recurrent neural
  networks.
\newblock In \emph{2017 IEEE 60th International Midwest Symposium on Circuits
  and Systems (MWSCAS)}, pages 1593--1596, 2017.
\newblock \doi{10.1109/MWSCAS.2017.8053242}.

\bibitem[Jouppi et~al.(2017)Jouppi, Young, Patil, Patterson, Agrawal, Bajwa,
  Bates, Bhatia, Boden, Borchers, Boyle, luc Cantin, Chao, Clark, Coriell,
  Daley, Dau, Dean, Gelb, Ghaemmaghami, Gottipati, Gulland, Hagmann, Ho,
  Hogberg, Hu, Hundt, Hurt, Ibarz, Jaffey, Jaworski, Kaplan, Khaitan, Koch,
  Kumar, Lacy, Laudon, Law, Le, Leary, Liu, Lucke, Lundin, MacKean, Maggiore,
  Mahony, Miller, Nagarajan, Narayanaswami, Ni, Nix, Norrie, Omernick,
  Penukonda, Phelps, Ross, Ross, Salek, Samadiani, Severn, Sizikov, Snelham,
  Souter, Steinberg, Swing, Tan, Thorson, Tian, Toma, Tuttle, Vasudevan,
  Walter, Wang, Wilcox, and Yoon]{1704.04760}
Norman~P. Jouppi, Cliff Young, Nishant Patil, David Patterson, Gaurav Agrawal,
  Raminder Bajwa, Sarah Bates, Suresh Bhatia, Nan Boden, Al~Borchers, Rick
  Boyle, Pierre luc Cantin, Clifford Chao, Chris Clark, Jeremy Coriell, Mike
  Daley, Matt Dau, Jeffrey Dean, Ben Gelb, Tara~Vazir Ghaemmaghami, Rajendra
  Gottipati, William Gulland, Robert Hagmann, C.~Richard Ho, Doug Hogberg, John
  Hu, Robert Hundt, Dan Hurt, Julian Ibarz, Aaron Jaffey, Alek Jaworski,
  Alexander Kaplan, Harshit Khaitan, Andy Koch, Naveen Kumar, Steve Lacy, James
  Laudon, James Law, Diemthu Le, Chris Leary, Zhuyuan Liu, Kyle Lucke, Alan
  Lundin, Gordon MacKean, Adriana Maggiore, Maire Mahony, Kieran Miller, Rahul
  Nagarajan, Ravi Narayanaswami, Ray Ni, Kathy Nix, Thomas Norrie, Mark
  Omernick, Narayana Penukonda, Andy Phelps, Jonathan Ross, Matt Ross, Amir
  Salek, Emad Samadiani, Chris Severn, Gregory Sizikov, Matthew Snelham, Jed
  Souter, Dan Steinberg, Andy Swing, Mercedes Tan, Gregory Thorson, Bo~Tian,
  Horia Toma, Erick Tuttle, Vijay Vasudevan, Richard Walter, Walter Wang, Eric
  Wilcox, and Doe~Hyun Yoon.
\newblock In-datacenter performance analysis of a tensor processing unit.
\newblock In \emph{Proceedings of the 44th Annual International Symposium on
  Computer Architecture}, ISCA '17, page 1–12, 2017.
\newblock \doi{10.1145/3079856.3080246}.

\bibitem[Matula(1987)]{Matula87}
David~W. Matula.
\newblock Determining edge connectivity in 0(nm).
\newblock In \emph{28th Annual Symposium on Foundations of Computer Science
  (sfcs 1987)}, pages 249--251, 1987.
\newblock \doi{10.1109/SFCS.1987.19}.

\bibitem[Miranda and Zuben(2015)]{1511.02954}
Conrado~Silva Miranda and Fernando Jos{\'{e}}~Von Zuben.
\newblock Reducing the training time of neural networks by partitioning, 2015.
\newblock URL \url{https://arxiv.org/abs/1511.02954}.

\bibitem[Mohammadi et~al.(2021)Mohammadi, Fathy, and
  Sabokrou]{DBLP:journals/corr/abs-2103-01739}
Bahram Mohammadi, Mahmood Fathy, and Mohammad Sabokrou.
\newblock Image/video deep anomaly detection: {A} survey, 2021.
\newblock URL \url{https://arxiv.org/abs/2103.01739}.

\bibitem[Schmiedl(2017)]{DBLP:journals/dcg/Schmiedl17}
Felix Schmiedl.
\newblock Computational aspects of the gromov-hausdorff distance and its
  application in non-rigid shape matching.
\newblock \emph{Discrete {\&} Computational Geometry}, 57\penalty0
  (4):\penalty0 854--880, 2017.
\newblock \doi{10.1007/s00454-017-9889-4}.

\bibitem[Simonyan and Zisserman(2015)]{SimonyanZ2014}
Karen Simonyan and Andrew Zisserman.
\newblock Very deep convolutional networks for large-scale image recognition.
\newblock In Yoshua Bengio and Yann LeCun, editors, \emph{3rd International
  Conference on Learning Representations, {ICLR} 2015}, 2015.
\newblock URL \url{http://arxiv.org/abs/1409.1556}.

\bibitem[Szegedy et~al.(2015)Szegedy, Liu, Jia, Sermanet, Reed, Anguelov,
  Erhan, Vanhoucke, and Rabinovich]{SzegedyLJSRAEVR2015}
Christian Szegedy, Wei Liu, Yangqing Jia, Pierre Sermanet, Scott Reed, Dragomir
  Anguelov, Dumitru Erhan, Vincent Vanhoucke, and Andrew Rabinovich.
\newblock Going deeper with convolutions.
\newblock In \emph{2015 IEEE Conference on Computer Vision and Pattern
  Recognition (CVPR)}, pages 1--9, 2015.
\newblock \doi{10.1109/CVPR.2015.7298594}.

\bibitem[Tan et~al.(2020)Tan, Pang, and Le]{DBLP:journals/corr/abs-1911-09070}
Mingxing Tan, Ruoming Pang, and Quoc~V. Le.
\newblock Efficientdet: Scalable and efficient object detection.
\newblock In \emph{2020 IEEE/CVF Conference on Computer Vision and Pattern
  Recognition (CVPR)}, pages 10778--10787, 2020.
\newblock \doi{10.1109/CVPR42600.2020.01079}.

\bibitem[Toneva et~al.(2019)Toneva, Sordoni, Combes, Trischler, Bengio, and
  Gordon]{Forgetting}
Mariya Toneva, Alessandro Sordoni, Remi Tachet~des Combes, Adam Trischler,
  Yoshua Bengio, and Geoffrey~J Gordon.
\newblock An empirical study of example forgetting during deep neural network
  learning.
\newblock In \emph{ICLR}, 2019.
\newblock URL \url{https://openreview.net/forum?id=BJlxm30cKm}.

\bibitem[Wang et~al.(2018)Wang, Huan, and Li]{8576015}
Tianyang Wang, Jun Huan, and Bo~Li.
\newblock Data dropout: Optimizing training data for convolutional neural
  networks.
\newblock In \emph{2018 IEEE 30th International Conference on Tools with
  Artificial Intelligence (ICTAI)}, pages 39--46, 2018.
\newblock \doi{10.1109/ICTAI.2018.00017}.

\bibitem[Wang et~al.(2020)Wang, Zhu, Dong, He, and
  Huang]{Wang_Zhu_Dong_He_Huang_2020}
Zifeng Wang, Hong Zhu, Zhenhua Dong, Xiuqiang He, and Shao-Lun Huang.
\newblock Less is better: Unweighted data subsampling via influence function.
\newblock \emph{Proceedings of the AAAI Conference on Artificial Intelligence},
  34\penalty0 (04):\penalty0 6340--6347, Apr. 2020.
\newblock \doi{10.1609/aaai.v34i04.6103}.
\newblock URL \url{https://ojs.aaai.org/index.php/AAAI/article/view/6103}.

\bibitem[Xiao et~al.(2017)Xiao, Zhu, Liu, and Zhang]{DBLP:conf/ijcai/XiaoZLZ17}
Tong Xiao, Jingbo Zhu, Tongran Liu, and Chunliang Zhang.
\newblock Fast parallel training of neural language models.
\newblock In \emph{Proceedings of the Twenty-Sixth International Joint
  Conference on Artificial Intelligence, {IJCAI} 2017, Melbourne, Australia,
  August 19-25, 2017}, pages 4193--4199, 2017.
\newblock \doi{10.24963/ijcai.2017/586}.

\bibitem[You and Xu(2014)]{DBLP:conf/iscslp/YouX14}
Zhao You and Bo~Xu.
\newblock Improving training time of deep neural networkwith asynchronous
  averaged stochastic gradient descent.
\newblock In \emph{The 9th International Symposium on Chinese Spoken Language
  Processing}, pages 446--449, 2014.
\newblock \doi{10.1109/ISCSLP.2014.6936596}.

\bibitem[Yuan et~al.(2019)Yuan, He, Zhu, and Li]{1712.07107}
Xiaoyong Yuan, Pan He, Qile Zhu, and Xiaolin Li.
\newblock Adversarial examples: Attacks and defenses for deep learning.
\newblock \emph{IEEE Transactions on Neural Networks and Learning Systems},
  30\penalty0 (9):\penalty0 2805--2824, 2019.
\newblock \doi{10.1109/TNNLS.2018.2886017}.

\end{thebibliography}
%\bibliographystyle{plainnat}


\end{document}